\documentclass[11pt]{article}

\usepackage[T1]{fontenc}
\usepackage[utf8]{inputenc}
\usepackage{lmodern}
\usepackage{graphicx}
\usepackage{fullpage}
\usepackage{amssymb,amsmath}
\usepackage{bm}
\usepackage{natbib}

\DeclareMathOperator*{\argmin}{argmin}
\DeclareMathOperator*{\argmax}{argmax}

\title{Expectation Propagation
\footnote{
These are notes from lectures by Manfred Opper given at the autumn
school “Statistical Physics, Optimization, Inference, and Message-Passing
Algorithms”, that took place at Les Houches, France, from September 30th to
October 11th 2013. The school was organized by Florent Krzakala from UPMC \&
ENS Paris, Federico Ricci-Tersenghi from La Sapienza Roma, Lenka Zdeborová
from CEA Saclay \& CNRS, and Riccardo Zecchina from Politecnico Torino.}
}
\author{Jack Raymond, Andre Manoel, Manfred Opper}
\date{}

\begin{document}
\maketitle


\abstract{Variational inference is a powerful concept that underlies many
    iterative approximation algorithms; expectation propagation, mean-field
    methods and belief propagations were all central themes at the school
    that can be perceived from this unifying framework. The lectures of
    Manfred Opper introduce the archetypal example of Expectation
    Propagation, before establishing the connection with the other
    approximation methods.  Corrections by expansion about the expectation
    propagation are then explained. Finally some advanced inference topics
    and applications are explored in the final sections.}

\tableofcontents

\newcommand{\dbar}{\,|\kern-0.1em|\,}
\graphicspath{{figures/}}

\section{Introduction}
These notes were prepared in October and November 2013. The authors have
followed the structure of the original presentation and the second lecture
begins at section \ref{sec:GibbsFreeEnergy}. 
All figures and topics from the original presentation slides are included
and the style of the lecture was followed: For brevity and continuity we do
not attempt a detailed description of every application, providing instead
references; We try to minimize backward references -- repeated concepts such
as the {\em tilted distribution} and {\em Gaussian prior} are redefined
locally in most cases; Finally, so as to focus on the important parameter
that is the subject of approximation, we frequently omit from the likelihood
or posterior the data argument.

Wherever possible we provide the references to the original work discussed,
we have also included at our discretion some additional references, in
particular the book by Wainwright and Jordan provides a very elegant
introduction to the topics discussed in the first lecture, and beginning of
the second lecture (\citealp{wainwright_graphical_2008}).

\section{Approximate variational inference}
We are interested in computing the statistics of hidden variables $\bm{x} =
(x_1, x_2, \dots, x_n)$ given the observed data $\bm{y} = (y_1, y_2, \dots,
y_m)$ and a generative model relating $\bm{x}$ to $\bm{y}$, specified by the
joint distribution $p (\bm{x}, \bm{y})$. For this purpose, it is convenient
to work with the posterior distribution $p (\bm{x} | \bm{y})$, which from
the Bayes theorem may be written as

\begin{equation}
	p (\bm{x} | \bm{y}) = \frac{p (\bm{x}, \bm{y})}{p (\bm{y})}
\end{equation}
and then focus in on one or more of the following
tasks (\citealp{wainwright_graphical_2008}):

\begin{itemize} 
\item determining the mode of the posterior, $\argmax_{\bm{x}} p(\bm{x} |
    \bm{y})$, which gives us the most probable assignment of $\bm{x}$ for an
    instance of $\bm{y}$;
\item obtaining the marginals $p(x_i | \bm{y}) = \int d\bm{x}_{\sim i} \, p
    (\bm{x} | \bm{y})$, which allows us to compute $x_i$ moments and
    estimators ($\bm{x}_{\sim i}$ denotes the set of variables except
    $x_i$);
\item computing the model evidence $p (\bm{y}) = \int d\bm{x} \, p(\bm{x},
    \bm{y})$, which may be used for model selection as well for computing
    the moments of sufficient statistics in exponential families, where $p
    (\bm{y})$ is also known as partition function.
\end{itemize}

Both the marginals and evidence computations involve high dimensional sums
or integrals and are often intractable. A \emph{variational
approximation} (\citealp{bishop_pattern_2007,
barber_bayesian_2011,murphy_probabilistic}) consists in replacing $p (\bm{x}
| \bm{y})$ by $q (\bm{x}) \in \mathcal{F}$, where $\mathcal{F}$ is a
tractable\footnote{for which computing the marginals or the evidence may be
performed exactly and in polynomial time in $n$.} family of distributions,
so as to minimize any desired measure between $p(\bm{x} | \bm{y})$ and $q
(\bm{x})$. In particular, KL variational approximations pick $q (\bm{x})$ so
as to minimize the following Kullback-Leibler divergence

\begin{align}
    \operatorname{KL} \left[ q (\bm{x}) \dbar p (\bm{x} | \bm{y}) \right] &=
        \int d\bm{x} \, q (\bm{x}) \log \frac{q (\bm{x})}{p (\bm{x} | \bm{y})}
        \notag \\
    &= \underbrace{-\mathbb{E}_q \log p (\bm{x}, \bm{y})}_{E [q]} +
        \underbrace{\int d\bm{x} \, q (\bm{x}) \log q (\bm{x})}_{-S[q]} + \log p
        (\bm{y}) \geq 0 \label{eq:KL1}
\end{align}

Such approximations correspond to the so-called mean-field theories of
statistical physics (\citealp{opper_advanced_2001}), where the quantities $E[q]$
and $S [q]$ assume the roles of  energy and entropy, respectively. By
introducing a variational free energy $F[q] = E[q] - S[q]$, we recover the
known bound $-\log p(\bm{y}) \leq F[q]$.

\subsection{Different families/approximations}

The next step is to pick a family of distributions. For a fully factorized
distribution $q (\bm{x}) = \frac{1}{\mathcal{Z}} \prod_{i = 1}^n q_i (x_i)$,
the ${\{q_i^\ast\}}$ that minimize $F[q]$ are given by 

\begin{equation}
    q_j^\ast (x_j) = \frac{1}{z_j} \exp \left\{ -\mathbb{E}_{q^{\ast}_{\sim
        j}} \log p(\bm{x}, \bm{y}) \right\}
\end{equation}
where $\mathbb{E}_{q^\ast_{\sim j}}$ denotes average over all $\{q_i^\ast\}$
distributions except $q_j^\ast$. This choice defines the \emph{naive
mean-field} approximation. It may be applied to both discrete and continuous
models; in particular, for the Ising model $-\mathbb{E}_{q_{\sim j}} \log
p(\bm{x}, \bm{y}) = x_j \left( \sum_{k \in \mathcal{N} (j) } J_{jk} \langle
    x_k \rangle + B_j \right)$, and the familiar expression 

\begin{equation}
    \langle x_j \rangle = \tanh \left( B_j + \sum_{k } J_{jk} \langle x_k
        \rangle \right)
    \label{eq:nmf_ising}
\end{equation}
is recovered.  The weak point in this approach is that it neglects
dependencies between variables. Linear response corrections are possible,
and can yield non-zero estimates of correlations correct at leading order,
but have themselves some weaknesses. 

A Gaussian approximation, $q (\bm{x}) \propto \exp \left\{-\frac{1}{2}
(\bm{x} - \bm{\mu})^T \Sigma^{-1} (\bm{x} - \bm{\mu}) \right\}$,
equivalently leads to 

\begin{equation}
    \mathbb{E}_{q^\ast} \left\{ \frac{\partial \log p (\bm{x},
        \bm{y})}{\partial x_i} \right\} = 0 \kern3.5em
    \mathbb{E}_{q^\ast} \left\{ \frac{\partial^2 \log p (\bm{x},
        \bm{y})}{\partial x_i \partial x_j} \right\} =
        \left(\Sigma^{-1}\right)_{ij}
\end{equation}
however discrete models may not be considered in this case, since for the KL
divergence (\ref{eq:KL1}) to be well defined the support of the distribution
$p(x|y)$ must always include that of $q(x)$. 

Slightly different approaches lead to other popular approximations. For
instance, for $q (\bm{x}) = \frac{\prod_{ij} q_{ij} (x_i, x_j)}{\prod
q_i^{d_i - 1} (x_i)}$, minimizing $F [q]$ while taking into account
normalization and consistency constraints leads to the loopy belief
propagation algorithm (\citealp{yedidia_bethe_2000}).

\section{Expectation propagation}
We could also consider the minimization of the reverse KL divergence,
$\operatorname{KL} [p (\bm{x} | \bm{y})  \dbar q (\bm {y})]$, instead of
$\operatorname{KL} [q (\bm {y}) \dbar p (\bm{x} | \bm{y})]$; this problem is
however intractable in general, since we would have to compute averages with
respect to $p (\bm{x} | \bm{y})$. Let us consider $q (\bm{x})$ distributions
belonging to the exponential family

\begin{equation}
 q (\bm{x}) = h (\bm{x}) \exp \left\{ \bm{\theta}^T \bm{\phi} (\bm{x}) + g(\bm{\theta}) \right\}
\end{equation}
of natural parameters $\bm{\theta}$ and sufficient statistics $\bm{\phi}
(\bm{x})$; for example, a multivariate normal distribution has $\bm{\theta}
= (\Sigma^{-1} \bm{\mu}, -\frac{1}{2} \Sigma^{-1})$ and $\bm{\phi} (\bm{x})
= (\bm{x}, \bm{x} \bm{x}^T)$. For this family, the minimization of the
reverse KL divergence results in

\begin{align}
    \nabla_{\bm{\theta}} \operatorname{KL} [p (\bm{x} | \bm{y})  \dbar q
        (\bm {y})] &= -\int d\bm{x} \, p (\bm{x} | \bm{y}) \bm{\phi}(\bm{x}) -
        \underbrace{\int d\bm{x} \, p (\bm{x} | \bm{y}) \nabla_{\bm{\theta}}
        g(\bm{\theta})}_{\mathbb{E}_p \nabla_{\bm{\theta}} g(\bm{\theta}) =
        -\mathbb{E}_q \bm{\phi} (\bm{x})} \notag \\ &= -\mathbb{E}_p
            \bm{\phi} (\bm{x}) + \mathbb{E}_q \bm{\phi} (\bm{x}) = 0
\end{align}
that is, in moment matching of the sufficient statistics $\bm{\phi}
(\bm{x})$; the problem will thus be tractable whenever the moments with
respect to $p (\bm{x} | \bm{y})$ may be efficiently computed.

\subsection{Assumed density filtering}

If we assume observed data to arrive sequentially, $\mathcal{D}_{t+1} =
\left\{y_1, y_2, \dots, y_{t+1} \right\}$, we may incorporate the new
measurement to the posterior, at each step, by means of the Bayes rule $p
(\bm{x} | \mathcal{D}_{t+1}) \propto p (y_{t + 1} | \bm{x}) \, p (\bm{x} |
\mathcal{D}_t)$.

In the assumed density filtering algorithm, the posterior $p (\bm{x} |
\mathcal{D}_{t + 1})$ is replaced, at each step, by the distribution
$q_{\theta} (\bm{x}) \in \mathcal{F}$ which minimizes $\operatorname{KL} [p
(\bm{x} | \mathcal{D}_{t +1}) \dbar q_{\theta} (\bm {x})]$. An iteration of
the algorithm thus consists in 

\begin{description}
\item[Initialize] by setting $q_\theta^{(0)} (\bm{x}) = p_0 (\bm{x})$.

\item[Update] the posterior

\begin{equation}
p (\bm{x} | \mathcal{D}_{t+1}) = \frac{p (y_{t + 1} | \bm{x}) \,
    q_\theta^{(t)} (\bm{x})}{\int d\bm{x} \, p (y_{t + 1} | \bm{x}) \,
    q_\theta^{(t)} (\bm{x})}
\label{eq:upd_adf}
\end{equation}

\item[Project] it back to $\mathcal{F}$

\begin{equation}
q_{\theta}^{(t + 1)} (\bm{x}) = \argmin_{q_{\theta} \in \mathcal{F}}
    \operatorname{KL} [p (\bm{x} | \mathcal{D}_{t +1}) \dbar q_{\theta} (\bm
    {x})]
\label{eq:proj_adf}
\end{equation}
\end{description}
 
The key premise is that by minimizing the reverse KL divergence while
including the contribution of a single factor of the (intractable)
likelihood, we ensure the problem remains tractable.  For a simple example,
consider the Bayesian classifier given by $y_t = \operatorname{sgn} \left(
h_{\bm{w}} (s_t) \equiv \sum_j w_j \psi_j (s_t) \right)$, leading to a
probit likelihood 

\begin{equation}
p (y_t | \bm{w}, s_t) = \frac{1}{2} + \frac{1}{\sqrt{2\pi}} \int_0^{y
h_{\bm{w}} (s_t)} e^{-\frac{u^2}{2}} \, du
\end{equation}

\begin{figure}
	\centering
	\includegraphics[width=0.5\textwidth]{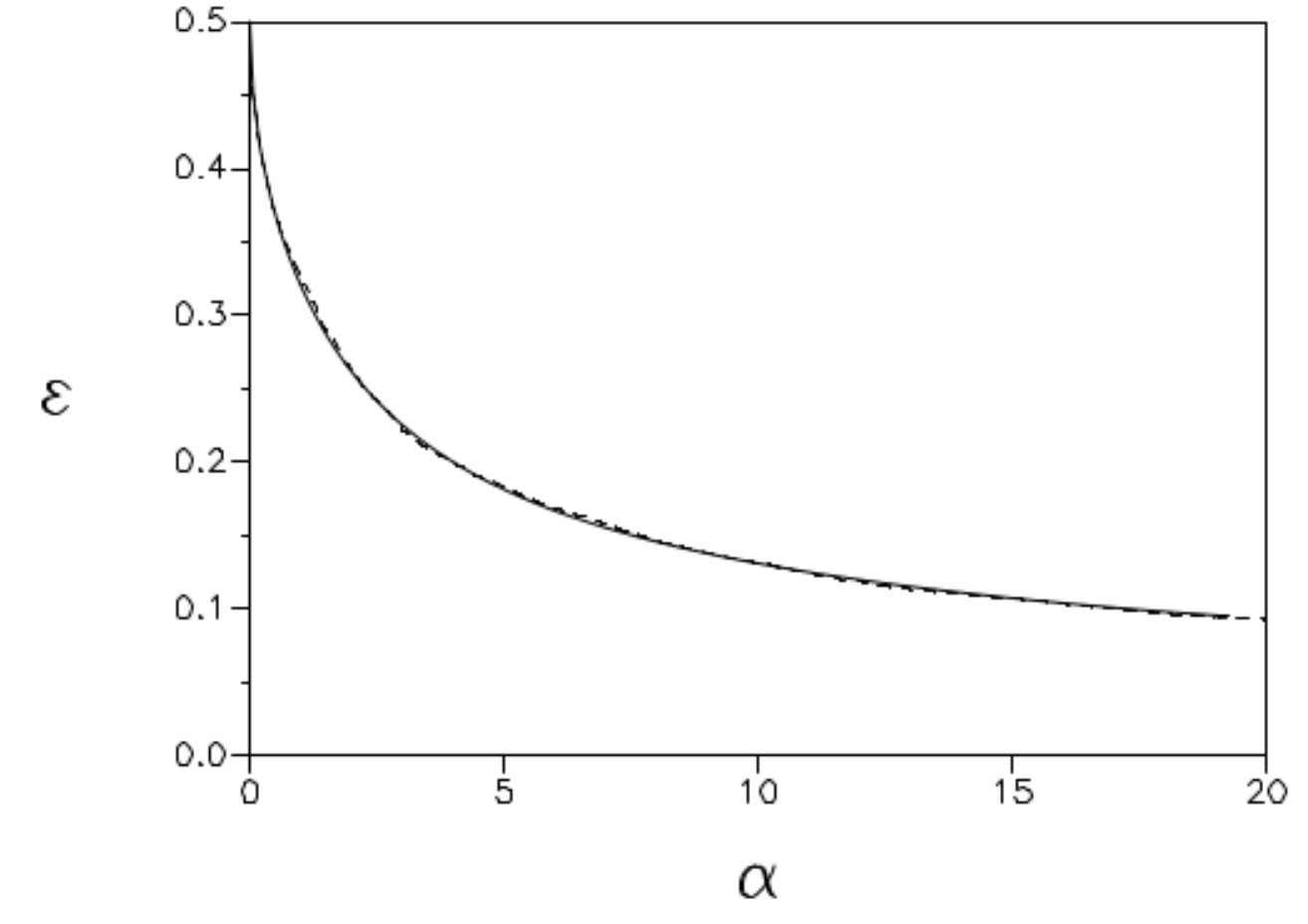}
    \caption{Generalization error $\varepsilon$ for a Bayesian classifier
        with $n = 50$ in terms of $\alpha = \frac{m}{n}$. The solid line gives
        the performance of the ADF algorithm, and the dashed one the Bayes optimal
        performance obtained via replica calculations. Figure extracted from
        \citealp{opper_1989}.}
    \label{fig:adf_classif}
\end{figure}

The update rule (\ref{eq:upd_adf}) in this case yields

\begin{equation}
	p(\bm{w} | y_{t+1}, s_{t+1}) \propto p(y_{t+1} | \bm{w}, s_{t+1}) \, q_\theta (\bm{w})
\end{equation}
and for a Gaussian prior over the weights $p_0 (\bm{w}) \propto \exp \left(
-\frac{1}{2} \sum_j w_j^2 \right)$ and a parametric approximation
$q_{\theta} (\bm{w}) = \mathcal{N} (\bm{\bar{w}}, C)$, the minimization
procedure (\ref{eq:proj_adf}) is easily accomplished, since $p(y_{t+1} |
\bm{w}, s_{t+1})$ depends on a single Gaussian integral. Results obtained by
applying the ADF algorithm to this model are presented in figure
\ref{fig:adf_classif}.

The weakness in ADF lies in the fact that the final result depends on the
order of presentation of data. Trying to fix this issue leads us to the
expectation propagation (EP) algorithm.

\subsection{The EP algorithm}
Given all the $m$ measurements $y_\mu$, the posterior may be written as

\begin{equation}
    p(\bm{x} | \bm{y}) = \frac{p_0 (\bm{x}) \prod_{\mu = 1}^{m} p (y_\mu |
    \bm{x})}{\int d\bm{x} \, p_0 (\bm{x}) \prod_{\mu = 1}^{m} p (y_\mu |
    \bm{x})}  = \frac{1}{\mathcal{Z}} f_0 (\bm{x}) \prod_{\mu = 1}^m f_\mu
    (\bm{x})
\end{equation}
where we assume the prior $f_0 (\bm{x})$ to be a member of the exponential
family. The expectation propagation (EP)
algorithm (\citealp{minka_expectation_2001}) consists in determining a tractable
approximation $q (\bm{x}) = \frac{1}{\mathcal{Z}} f_0 (\bm{x}) \prod_{\mu =
1}^m g_\mu (\bm{x})$ to $p (\bm{x} | \bm{y})$ by repeatedly performing the
ADF update (\ref{eq:upd_adf}) as if each measurement has never been
presented before, effectively by removing its contribution from the current
approximation. The whole procedure consists in 

\begin{description}
\item[Initialize] by setting $q (\bm{x}) = f_0 (\bm{x})$ and $g_\mu (\bm{x}) = 1$, $\mu \in \{1, 2, \dots, m\}$.

\item[Repeat] until convergence --- pick $\mu \in {1, 2, \dots, m}$ uniformly at random and 

\begin{description}
\item[Remove] $g_\mu (\bm{x})$ from $q (\bm{x})$, i.e., build $q_{\sim \mu} (\bm{x}) \propto \frac{q (\bm{x})}{g_\mu (\bm{x})}$.

\item[Update] the \emph{tilted} distribution $q_\mu (\bm{x}) = f_\mu (\bm{x}) q_{\sim \mu} (\bm{x})$.

Note that $q_\mu (\bm{x}) \propto \frac{f_\mu (\bm{x})}{g_\mu (\bm{x})} \, q
(\bm{x})$ replaces the approximating likelihood $g_\mu (\bm{x})$ by the
exact one $f_\mu (\bm{x})$, thus taking $q_\mu (\bm{x})$ outside of the
exponential family.

\item[Project] $q (\bm{x})$ back to the exponential family

\begin{equation}
	q^{\text{new}} (\bm{x}) = \argmin_{q \in \mathcal{E}} \operatorname{KL} \left[ q_\mu (\bm{x}) \dbar q (\bm{x}) \right]
\end{equation}

\item[Refine] the approximated likelihood terms

\begin{equation}
    g_\mu^{\text{new}} (\bm{x}) \propto \frac{q^{\text{new}}
        (\bm{x})}{q_{\sim \mu} (\bm{x})} \propto \frac{q^{\text{new}}
        (\bm{x})}{q (\bm{x})} g_\mu (\bm{x}) \qquad \text{for } \mu \in \{1, 2,
        \dots, m\}.
\end{equation}
\end{description}

\item[At convergence] obtain tractable approximation from $q (\bm{x}) = f_0 (\bm{x}) \prod_{\mu = 1}^m f_\mu (\bm{x})$.

    The approximated posterior $q (\bm{x})$ and the tilted distributions
    $q_{\mu} (\bm{x}) = \frac{f_\mu (\bm{x})}{g_\mu (\bm{x})}$ will have
    matching moments, $\mathbb{E}_q \phi (\bm{x}) = \mathbb{E}_{q_\mu}
    \phi(\bm{x})$ for $\mu \in \{1, 2, \dots, m\}$.
\end{description}

We assume here that the $\{g_\mu (\bm{x})\}$ belong to the exponential
family; in order to proceed, one needs to choose a distribution from this
family, i.e., to specify how $q (\bm{x})$ is to be factorized. Let us
exemplify the procedure by considering a Gaussian latent variable model with
$m = n$, $p (\bm{x} | \bm{y}) \propto e^{ -\frac{1}{2} \bm{x}^T K \bm{x}}
\prod_{i = 1}^n p (y_i | \bm{x})$. We will pick Gaussian $g_i (x_i) \propto
\exp (-\frac{1}{2} \Lambda_i x_i^2 + \gamma_i x_i)$ so that

\begin{equation}
	q (\bm{x}) \propto \exp \left\{ -\frac{1}{2} \bm{x}^T K \bm{x} - \frac{1}{2} \sum_{i = 1}^n \Lambda_i x_i^2 + \bm{\gamma}^T \bm{x} \right\}
 \label{eq:gaussianApprox}
\end{equation}

The steps described above amount to, at each iteration, removing the
$\{\Lambda_i, \gamma_i\}$ terms for a given $i$; computing the marginal $q_i
(x_i) = \int d\bm{x}_{\sim i} \, p(y_i | \bm{x}) q_{\sim i} (\bm{x})$ and,
from it, the moments $\mathbb{E}_{q_i} x_i, \mathbb{E}_{q_i} x_i^2$; and
subsequently updating $\{\Lambda_i, \gamma_i\}$ so as to have $\mathbb{E}_q
x_i = \mathbb{E}_{q_i} x_i$ and $\mathbb{E}_q x_i^2 = \mathbb{E}_{q_i}
x_i^2$.

Empirically, it is verified that EP is a fast algorithm, however its
convergence is not guaranteed. It has the advantage of being applicable to
both discrete and continuous variable models, in particular providing
remarkable results for Gaussian latent variable models.

\subsection{Relation to BP}

It is not hard to show (see e.g. \citealp{murphy_probabilistic}) that, by
applying the algorithm above to any distribution $p (\bm{x} | \bm{y})
\propto f_0 (\bm{x}) \prod_{\mu = 1}^m f_\mu (\bm{x})$ with $g_\mu (\bm{x})
\propto \prod_{i \in \mathcal{N} (\mu)} h_{\mu i} (x_i)$, one recovers the
loopy belief propagation algorithm (\citealp{minka_expectation_2001}), where
$\mathcal{N} (\mu)$ are the set of variables coupled through $\mu$. The
approximated distribution is given by

\begin{equation}
	q (\bm{x}) \propto f_0 (\bm{x}) \prod_{\mu = 1}^m \prod_{i \in \mathcal{N} (\mu)} h_{\mu i} (x_i)
    \label{eq:bp_node}
\end{equation}
e.g., on a pairwise graphical model, one would be replacing likelihoods of
the form $f_{ij} (x_i, x_j) = \exp (J_{ij} x_i x_j)$ with $g_{ij} (x_i, x_j)
\propto \exp (\lambda_{ij} (x_i) + \lambda_{ji} (x_j))$. 

By defining $q^{(i)} (x_i) \propto \prod_{\mu \in \mathcal{N} (i)} h_{\mu i}
(x_i)$, the removal step for a given $\mu$ leads to

\begin{equation}
    q_{\sim \mu}^{(i)} (x_i) \propto \frac{q^{(i)} (x_i)}{h_{\mu i} (x_i)}
        \propto \prod_{\nu \in \mathcal{N} (i) \backslash \mu} h_{\nu i} (x_i)
        \qquad \text{for } i \in \mathcal{N} (\mu)
\end{equation}
while the projection step matches the marginals $q (\bm{x})$ and $q_{\mu}
(\bm{x})  \propto f_\mu (\bm{x}) \prod_{i \in \mathcal{N} (\mu)}
q^{(i)}_{\sim \mu} (x_i)$, that is

\begin{equation}
    q^{(i)} (x_i) = \sum_{\bm{x}_{\sim i}} q_{\mu} (\bm{x}) \propto
    \sum_{\bm{x}_{\sim i}} f_{\mu} (\bm{x}) \prod_{i \in \mathcal{N} (\mu)}
    q_{\sim \mu}^{(i)} (x_i) \qquad \text{for } i \in \mathcal{N} (\mu)
    \label{eq:bp_marg}
\end{equation}
and at the refine step, finally

\begin{equation}
    h_{\mu i} (x_i) \propto \frac{q^{(i)} (x_i)}{q^{(i)}_{\sim \mu} (x_i)}
        \propto \sum_{\bm{x}_{\sim i}} f_{\mu} (\bm{x}) \prod_{j \in \mathcal{N}
        (\mu) \backslash i} q_{\sim \mu}^{(j)} (x_j)
    \label{eq:bp_fac}
\end{equation}

Comparing these to the BP equations on a factor
graph (\citealp{wainwright_graphical_2008}), we can see that (\ref{eq:bp_node})
and (\ref{eq:bp_fac}) give the messages from nodes to factors and factors to
nodes respectively, while (\ref{eq:bp_marg}) provides approximations to the
marginals.

\section{Adaptive TAP}
For models with pairwise interactions the naive mean-field approximation may
be improved by means of the linear response correction

\begin{equation}
    \frac{\partial \langle x_i\rangle}{\partial B_j} = \langle x_i x_j \rangle -
        \langle x_i \rangle \langle x_j \rangle
    \label{eq:tap_linresp}
\end{equation}
where we would compute $\frac{\partial \langle x_i\rangle}{\partial B_j}$
within the current approximation and then set the correlations to $\langle
x_i x_j \rangle = \langle x_i \rangle \langle x_j \rangle + \frac{\partial
\langle x_i\rangle}{\partial B_j}$. For many interesting families we can
calculate exactly the linear response estimate at leading order, which is
sufficient in certain mean-field settings. For the Ising model
(\ref{eq:nmf_ising}) with i.i.d. couplings that would lead to a new term in
the free energy, since now $\mathbb{E}_q J_{ij} x_i x_j = J_{ij} \langle x_i
\rangle \langle x_j \rangle + J_{ij}^2 (1 - \langle x_i \rangle^2) (1 -
\langle x_j \rangle^2)$; the new mean-field equations are given by

\begin{equation}
    \langle x_j \rangle = \tanh \left( B_j + \sum_{k \in \mathcal{N} (j)}
        J_{jk} \langle x_k \rangle - \langle x_j \rangle \sum_{k \in \mathcal{N}
        (j)} J_{jk}^2 (1 - \langle x_k \rangle^2 )\right)
    \label{eq:tap}
\end{equation}
which are known as the TAP equations in statistical physics, while the new
term in the equations is called the Onsager reaction term.  There are more
systematic ways of deriving such corrections; one of them is the Plefka
expansion and another the cavity approach discussed next.

\subsection{Cavity approach}

Let us consider a probability distribution over pairwise interactions

\begin{equation}
    p (\bm{x}) = \frac{1}{\mathcal{Z}} \prod_{j = 1}^n f_j (x_j) \exp \Bigg(
        \sum_{jk} J_{jk} x_j x_k  + \sum_j B_j x_j \Bigg) \propto f_i (x_i) \exp
        \Bigg\{ x_i \Big( B_i + \underbrace{\sum_{j \in \mathcal{N} (i)} J_{ij}
        x_j}_{h_i} \Big) \Bigg\} p (\bm{x}_{\sim i})
    \label{eq:pairwise}
\end{equation}
whereby introducing the \emph{cavity field} distribution

\begin{equation}
    p_{\sim i} (h_i) = \int d\bm{x}_{\sim i} \, \delta \left( h_i - \sum_{j
        \in \mathcal{N} (i)} J_{ij} x_j \right) p (\bm{x}_{\sim i})
\end{equation}
the marginal distributions may be written as $p_i (x_i) \propto f_i (x_i)
\int dh_i \, p_{\sim i} (h_i) \exp \{ x_i (B_i + h_i) \}$. Repeating the
procedure for the $p (\bm{x}_{\sim i})$ by writing them in terms of $h_{j
\in \mathcal{N} (i)}$ would lead to the belief propagation equations; we
will proceed however by considering the large connectivity limit, and apply
the central limit theorem according to which $p_{\sim i} (h_i)$ should be
well approximated by a Gaussian. By setting $p_{\sim i} (h_i) \propto \exp
\left\{ -\frac{(h_i - a_i)^2}{2 V_i} \right\}$, the marginals may be
computed from

\begin{equation}
    p_i (x_i) = \frac{1}{z_i} f_i (x_i) \exp \left\{ (B_i + a_i) x_i +
        \frac{V_i}{2} x_i^2 \right\}
    \label{eq:tap_marg}
\end{equation}

The $\{a_i\}$ can be determined using the identity
$\langle h_i \rangle \propto \int dx_i \, f_i (x_i) \int h_i \, dh_i \,
p_{\sim i} (h_i) \, \exp (x_i h_i) = a_i + V_i \langle x_i \rangle = \sum_{j
\in \mathcal{N} (i)} J_{ij} \langle x_j \rangle$, so that
 
\begin{equation}
	a_i = \sum_{j \in \mathcal{N} (i)} J_{ij} \langle x_j \rangle - V_i \langle x_i \rangle
    \label{eq:tap_avg}
\end{equation}

The $\{V_i\}$ are by definition, $V_i = \sum_{jk} J_{ij} J_{jk} \left(
\mathbb{E}_{p_{\sim i}} x_j x_k - \mathbb{E}_{p_{\sim i}} x_j \,
\mathbb{E}_{p_{\sim i}} x_k \right)$; for independently sampled couplings
and assuming that $\mathbb{E}_{p_{\sim i}} \sim \mathbb{E}_{p}$, one may
write

\begin{equation}
	V_i = \sum_{j \in \mathcal{N} (i)} J_{ij}^2 (1 - \langle x_j \rangle^2)
    \label{eq:tap_var}
\end{equation}
and by substituting (\ref{eq:tap_var}) into (\ref{eq:tap_avg}), we get $a_i
= \sum_{j \in \mathcal{N} (i)} J_{ij} \langle x_j \rangle - \langle x_i
\rangle \sum_{j \in N_i} J_{ij}^2 (1 - \langle x_j \rangle^2 )$. For $x_i
\in \left\{\pm 1 \right\}$, $\langle x_i \rangle = \tanh (B_i + a_i)$ thus
recovering the TAP equations (\ref{eq:tap}).

\subsection{Adaptive correction for general J ensembles}

\begin{figure}[ht]
	\centering
	\includegraphics[width=0.5\textwidth]{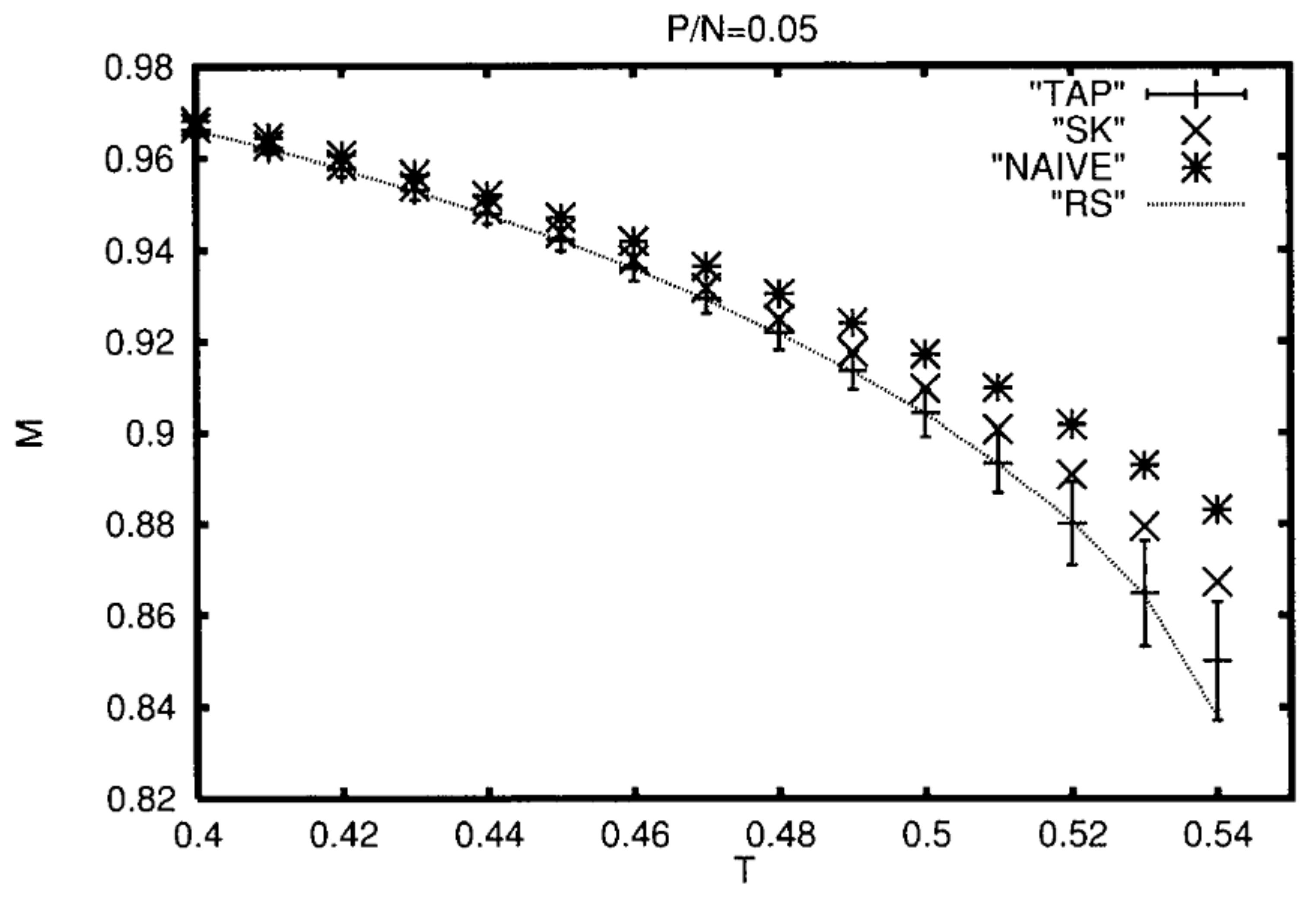}
    \caption{Comparison of the average overlap ($M$)
        in the Hopfield model obtained by means of iterating the naive mean
        field equations (NAIVE), the TAP equations with i.i.d. assumptions (SK) and
        the correct TAP equations (TAP) derived (see \citealp{kabashima_tap_2001}) for
        the statistics of the Hopfield couplings in a system with $N = 10000$. The
        Hopfield TAP equations successfully reproduce the replica-symmetric (RS)
        results, exactly for $N \to \infty$.  Figure extracted
        from \citealp{kabashima_tap_2001}.}
    \label{fig:hopfield}
\end{figure}

\begin{figure}
	\centering
	\includegraphics[width=0.5\textwidth]{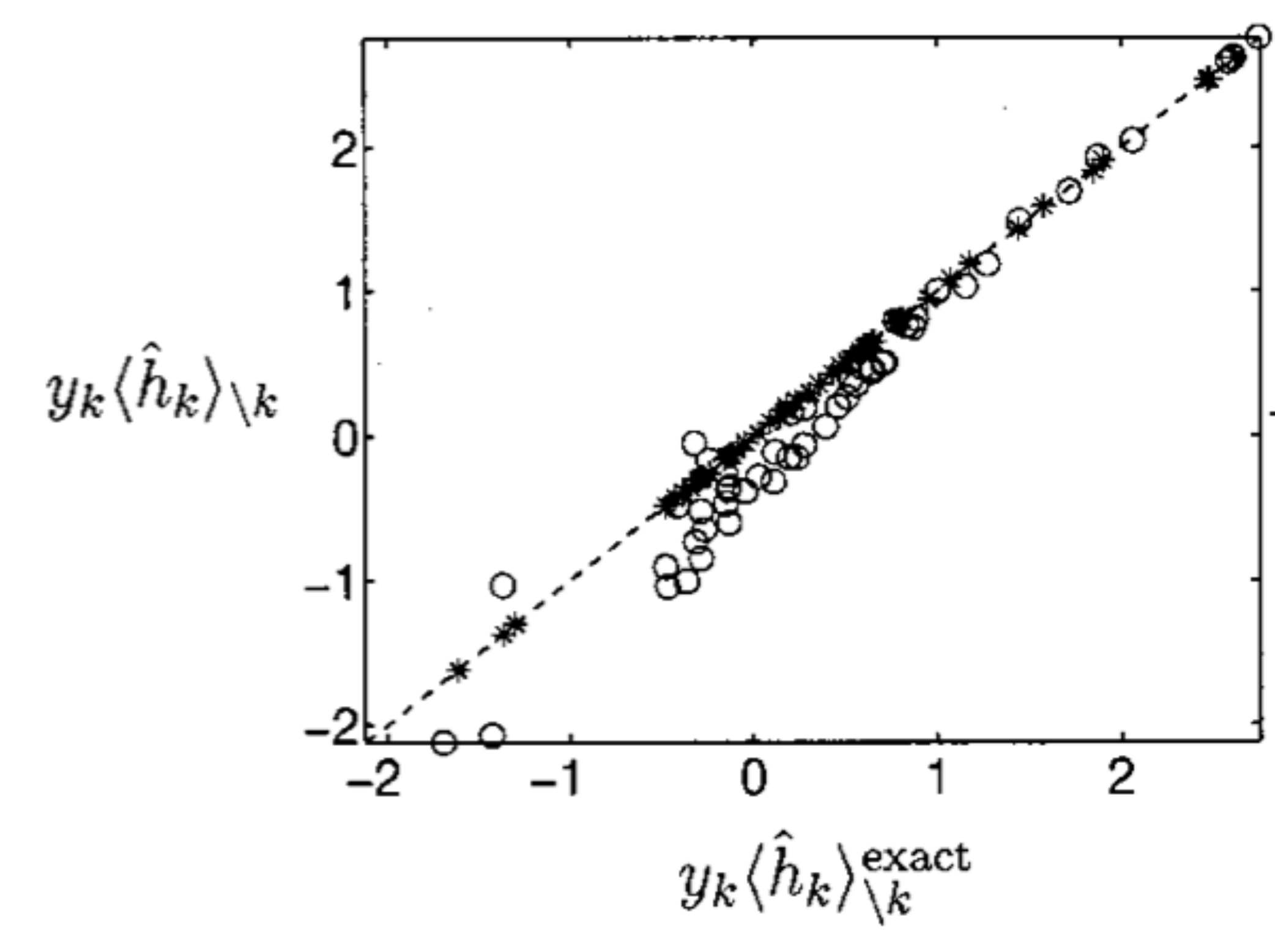}
    \caption{Testing TAP self-consistency in real data: the vertical axis
        gives the cavity field computed from the solution of the TAP equations,
        whereas the horizontal axis gives the \emph{exact} cavity field obtained by
        removing each sample from the training data and iterating the $m-1$
        remaining TAP equations. Stars and circles give the results for adaptive and
        conventional TAP, respectively. Figure extracted
        from \citealp{opper_adaptive_2001}.}
    \label{fig:tap_selfcons}
\end{figure}

In deriving (\ref{eq:tap_var}), we have assumed $J_{ij}$ and $J_{ik}$ to be
statistically independent, so that in the thermodynamical limit $n \to
\infty$, the off-diagonal terms $j \neq k$ vanish. While this assumption is
true for the SK model where $J_{ij} \sim \mathcal{N} (0, \frac{1}{n})$, it
breaks down when there are higher order correlations between the $\{ J_{ij}
\}$. For instance in Hopfield-like models, $J_{ij} = \sum_{p = 1}^{\alpha n}
\xi_{i}^{(p)} \xi_{j}^{(p)}$, these off-diagonal contributions do not vanish
(figure \ref{fig:hopfield}). Also when working with real data (figure
\ref{fig:tap_selfcons}), the distribution of the $\{J_{ij}\}$ is unknown,
and it is important to have a scheme which works independently of any such
assumptions.

While noting that $\langle x_i \rangle = \frac{\partial}{\partial B_i} \log
z_i \, (B_i, a_i, V_i)$, let us consider again the linear response relation,
(\ref{eq:tap_linresp}), from which a matrix $\chi$ of susceptibilities can
be defined

\begin{equation}
    \chi_{ij} = \frac{\partial \langle x_i \rangle}{\partial B_j} =
    \frac{\partial \langle x_i \rangle}{\partial B_i} \frac{\partial
    B_i}{\partial B_j} + \frac{\partial \langle x_i \rangle}{\partial a_i}
    \frac{\partial a_i}{\partial B_j} = \frac{\partial \langle x_i
    \rangle}{\partial B_i} \left( \delta_{ij} + \frac{\partial a_i}{\partial
    B_j} \right) 
\end{equation}
and an approximation that the $\{V_i\}$ are kept fixed has been made. By
further making use of (\ref{eq:tap_avg}), we get $\chi_{ij} = \frac{\partial
\langle x_i \rangle}{\partial B_i} \left\{ \delta_{ij} + \sum_{k \in
\mathcal{N} (i)} (J_{ik} + V_k \delta_{ik}) \, \chi_{ik} \right\}$, which
can be solved for $\chi$ to yield

\begin{equation}
    \chi_{ij} = \left[ (\Lambda - J)^{-1} \right]_{ij}
\end{equation}
where $\Lambda \equiv \operatorname{diag} \left( V_1 + \left(\frac{\partial
\langle x_1 \rangle}{\partial B_i} \right)^{-1}, \cdots, V_n +
\left(\frac{\partial \langle x_n \rangle}{\partial B_n} \right)^{-1}
\right)$ has been introduced.  The diagonal elements $\chi_{ii}$ may be
computed from this relation, but also from $\chi_{ii} =
\frac{\partial^2}{\partial B_i^2} \log z_i = \frac{1}{\Lambda_i - V_i}$; by
enforcing both solutions to be consistent, we obtain the following set of
equations

\begin{equation}
    \frac{1}{\Lambda_i - V_i} = \left[ (\Lambda - J)^{-1} \right]_{ii}
\end{equation}
which should be solved for the $V_i$, effectively replacing
(\ref{eq:tap_var}). Means and variances are then computed from

\begin{equation}
    \langle x_i \rangle = \frac{\partial}{\partial B_i} \, \log z_i (B_i,
    a_i^\ast, V_i^\ast) \qquad \langle x_i^2 \rangle - \langle x_i \rangle^2
    = \frac{\partial^2}{\partial B_i ^2} \, \log z_i (B_i, a_i^\ast,
    V_i^\ast)
\end{equation}
given the updated values $\{ a_i^\ast, V_i^\ast \}$. 

\subsection{Relation to EP}

Another way of deriving the adaptive correction comes from replacing the
$f_i (x_i)$ in (\ref{eq:tap_marg}) by a Gaussian $g_i (x_i) \propto
\exp(-\frac{1}{2} \Lambda_i x_i^2 - \gamma_i x_i)$, with $\Lambda_i,
\gamma_i$ chosen as to be consistent with $p_i (x_i)$ in the first two
moments $\langle x_i \rangle$ and $\langle x_i^2 \rangle$. Let

\begin{equation}
    z_i = \int dx_i \, f_i (x_i) \exp \left\{ a_i x_i + \frac{V_i}{2} x_i^2
    \right\} \qquad \tilde{z}_i = \int dx_i \, g_i (x_i) \exp \left\{ a_i
    x_i + \frac{V_i}{2} x_i^2 \right\}
\end{equation}
then by the moment matching requirement we have

\begin{align}
    \langle x_i \rangle &= \frac{d}{d a_i} \log z_i = \frac{d}{d a_i} \log
        \tilde{z}_i = \frac{\gamma_i + a_i}{\Lambda_i - V_i} \notag \\
    \langle x_i^2 \rangle - \langle x_i \rangle^2 &= \frac{d^2}{d a_i^2}
        \log z_i = \frac{d^2}{d a_i^2} \log \tilde{z}_i = \frac{1}{\Lambda_i -
        V_i}
    \label{eq:adatap1}
\end{align}

On the other hand, by direct computation of the moments of $p (\bm{x} |
\bm{y}) \propto \exp \left( \frac{1}{2} \bm{x}^T J \bm{x} \right) \prod_{i =
1}^n g_i (x_i)$

\begin{equation}
    \langle x_i \rangle = \left[ (\Lambda - J)^{-1} \bm{\gamma} \right]_i
        \qquad \langle x_i^2 \rangle - \langle x_i \rangle^2 = \left[ (\Lambda -
        J)^{-1} \right]_{ii}
    \label{eq:adatap2}
\end{equation}
and combining (\ref{eq:adatap1}) and (\ref{eq:adatap2}) leads us to the
relations obtained previously. This approach, which generalizes an idea by
Parisi and Potters, better ilustrates the similarity between adaptive TAP
and EP. In fact, running EP on a Gaussian latent model while using a
Gaussian approximation gives us the same fixed points as iterating the
equations above (\citealp{csato_tap_2001}).

\subsection{TAP approximation to free energy}
 
In order to write the TAP approximation to the free
energy (\citealp{opper_adaptive_2001}), we will introduce the $t$ variable which
mediates the strength between pairwise interactions

\begin{equation}
    p(\bm{x}) = \frac{1}{\mathcal{Z}_t} \exp\left( \frac{t}{2} \sum_{ij}
    J_{ij} x_i x_j \right) \prod_{i = 1}^n f_i (x_i)
\end{equation}

Since the TAP equations provide us the first and second moments of $\bm{x}$,
$m_i = \langle x_i \rangle$ and $M_i = \langle x_i^2 \rangle$, we work with
the Legendre transform of the free energy which keeps these quantities fixed

\begin{equation}
    G_t (\bm{m}, \bm{M}) = \min_{q} \left\{ \operatorname{KL} \left[ q (\bm{x}) \dbar p (\bm{x} | \bm{y}) \right] \, | \, \langle x_i \rangle_q = m_i, \langle x_i^2 \rangle_q = M_i \right\} - \log \mathcal{Z}_t
\end{equation}

We want to compute $G$ for $t = 1$, which may be done by using $G_1 = G_0 +
\int_0^1 dt \, \frac{\partial G_t}{\partial t}$; deriving $G_t$ with respect
to $t$ gives

\begin{equation}
    \frac{\partial G_t}{\partial t} (\bm{m}, \bm{M}) = -\frac{1}{2}
    \sum_{ij} J_{ij} \langle x_i x_j \rangle = -\frac{1}{2} \sum_{ij} J_{ij}
    m_i m_j - \frac{1}{2} \operatorname{Tr} \chi_t J
\end{equation}
then using the TAP approximation to $\chi$, $\chi_t = (\Lambda - t J)^{-1}$,
yields, after integration

\begin{equation}
    G_1 = G_0 -\frac{1}{2} \sum_{ij} J_{ij} m_i m_j +\frac{1}{2} \log \det
    (\Lambda - J) - \sum_i V_i (M_i - m_i^2) + \frac{1}{2} \sum_i \log (M_i
    - m_i^2)
    \label{eq:tap_free}
\end{equation}

\section{The Gibbs free energy}
\label{sec:GibbsFreeEnergy}
It has been shown (\citealp{opper_adaptive_2001, csato_tap_2001}) that the
Gibbs TAP free energy (\ref{eq:tap_free}) has an interpretation in terms of
a weighted sum of free energies for solvable models
\begin{equation}
    G_{\text{TAP}}(\mu) = G^{\text{Gauss}}(\mu) + G_0(\mu) -
    G^{\text{Gauss}}_0(\mu)\;, \label{eq:GibbsLect2} 
\end{equation}
These are the free energies for a global multivariate Gaussian approximation
$G^{\text{Gauss}}(\mu)$, a model which is factorized but contains non
Gaussian marginal distributions $G_0(\mu)$, and a part that is Gaussian and
factorized $G^{\text{Gauss}}_0(\mu)$.

Each Gibbs free energy is defined by the moments $(m,M)$ corresponding to
the single site statistics $\phi(x) = (x, -x^2/2)$, which are
\begin{equation}
    \mu = (\langle \phi(x_1)\rangle,\ldots,\langle \phi(x_N)\rangle\;.
\end{equation}

The Gibbs free energy is obtained from the Legendre transform of the
partition function for each of these models, given a set of fields $\lambda=
\{(\gamma_i,\Lambda_i): i =1 \ldots N\}$, we can write the partition
functions as
\begin{eqnarray}
    Z^{\text{Gauss}}(\lambda^{1}) = \int dx f_0(x) \exp\left(\sum_{i=1}^N
        (\lambda^{1}_i)^T\phi(x_i) \right)\;;\\
    Z_0(\lambda^{2}) = \int dx \prod_{i=1}^N f_i(x) \exp\left(\sum_{i=1}^N
        (\lambda^{2}_i)^T\phi(x_i) \right)\;;\\
    Z^{\text{Gauss}}_0(\lambda^{3}) = \int dx \exp\left(\sum_{i=1}^N
        (\lambda^{3}_i)^T\phi(x_i) \right)\;.
\end{eqnarray}
where $f_0(x) = \exp\left(\sum_{ij} x_i J_{ij} x_j\right)$.

The Legendre transform relates the representation in terms of fields, to the
representation in terms of moments. The Gibbs free energy is defined
\begin{equation}
    G(\mu) = \max_\lambda\lbrace -\ln Z(\lambda) + \lambda^T\mu \rbrace\;.
\end{equation}
Stationarity of the TAP free energy $\nabla_\mu G_{\text{TAP}}(\mu)=0$
implies that only two of the three fields $\lambda^{*}$ are independent
\begin{equation}
    -\log Z \approx -\log(Z_{\text{TAP}}) = -\log
    Z^{\text{Gauss}}(\lambda^1) - \log Z_0(\lambda^2) + \log
    Z^{\text{Gauss}}_0(\lambda^{1}+\lambda^{2}) \;.
\end{equation}

We find the TAP free energy by requiring stationarity of the right hand side
with respect to the fields.

\subsection{Double loop algorithms (minimization of non-convex Gibbs free
energies)}
Typically one can formulate the iterative algorithms discussed as minimizers
of some free energy, that are convergent under some sufficient set of
criteria; one pervasive criteria required for the success of these
algorithms (in general) is convexity of the free energy.  Unfortunately,
unlike the exact Gibbs free energy, approximate Gibbs free energies
$G_{approx}$ are often not convex, this is true of adaptive TAP for example.
However, in many cases, such as (\ref{eq:GibbsLect2}), there is a
decomposition of the form
\begin{equation}
    G_{approx}(\mu) = G_A(\mu) - G_B(\mu)
\end{equation}
where both $G_A$ and $G_B$ are convex. There exists a type of algorithm,
called a double loop algorithm, guaranteed to find local minima of
approximate free energies with this form (\citealp{Heskes:AICO}). This is
achieved by noting an upper bound $L(\mu)$ to the concave part $-G_B(\mu)$
defined
\begin{equation}
    L(\mu) = -G_B(\mu_{old}) - (\mu - \mu_{old})\nabla G_B(\mu_{old})
\end{equation}
The function $G_{vex}(\mu) = G_A(\mu) + L(\mu)$ is now convex and we can
apply standard methods. It can then be shown that by iteratively updating
$\mu_{old}$ we can converge to a minima, a standard implementation is
\begin{description}
    \item[Initialize] From random initial conditions $\mu_{old}$
    \item[Repeat] until convergence 
    \begin{description}
        \item[Update] $\mu_{new} = \argmin_\mu(G_{vex}(\mu))$
        \item[Update] $\mu_{old} = \mu_{new}$
    \end{description}
    \item[At convergence] $\mu=\mu_{new}$ minimizes the approximate $G_{approx}(\mu)$ free energy. 
\end{description}

Alternatives to double loop algorithms nevertheless continue to be popular,
despite problematic behaviour in some regimes. These include standard
message passing procedures such as Loopy Belief Propagation and Expectation
Propagation (using the recursive assumed density filtering procedure), and
other naive gradient descent methods. It is found that in many interesting
application domains these methods do converge. These algorithms are often
preferred owing to their significantly faster convergence.

\section{Improving and quantifying the accuracy of EP}
We have so far considered very simple approximations to $q$, involving
approximations to the marginal distributions by Gaussians. There are a
number of ways in which the approximations obtained might be improved. Two
ways are to consider more structured approximations (on trees rather than
single variables), and to consider expansions about the obtained solutions.
\begin{itemize}
    \item[1] For discrete random variables, one can consider extension to
        moment matching approximations, where consistency of diagonal
        statistics $\phi(x)=\{x,x^2\}$ is extended to pair statistics
        $\phi(x_i,x_j) = x_1 x_2$ (\citealp{Opper:ECFE,Minka:TSA}). 
    \item[2] We can expand about the approximation $q(x)$ to account at
        leading order for the structure ignored in the
        approximation (\citealp{Opper:IEP,Cseke:AMLGM}). We will consider two
        approximations based on expansions of the tilted distributions
        $q_n(x)$ about the EP approximation $q(x)$ (\citealp{Opper:IEP}). An
        expansion can proceed in either the difference of the two
        distributions, or in the higher order cumulants (first and second
        cumulants agree by definition).
\end{itemize} 

\subsection{The tree approximation}
In this case, and particularly for sparse graphs, we note that it is
possible to consider a more substantial part of the interaction structure
exactly, for example we can include pair statistics in the approximations
and require consistency of these moments.  The choice of additional pair
interactions has an important effect on the quality of the approximation
that will be obtained, we will want to include in the edge set the most
important interactions, and as many as possible such that the approximation
remains tractable.  A practical extension is to include all the pair and
vertex statistics defined by a tree within a Gaussian
approximation (\citealp{Opper:ECFE}). The tree can be chosen to cover the most
important correlations (by some practical computable criteria). For example
it can be chosen as a maximum weight spanning tree, with weights given by
the absolute values of the couplings.

Consider a spanning tree $\mathcal{T}$ that includes all the variables
$\{n\}$, and a set of edges $\{(m,n)\}$. With respect to the
tree\footnote{For simplicity we assume a single connected component,
otherwise the notation applies for a collection of trees.}, each variable
$n$ is said to have connectivity $d_n$. Assuming the probability
distribution is described by a Gaussian part, and a product of single
variable distributions, then it can be rewritten as
\begin{equation}
    p(x) = \frac{1}{Z} \exp\left(- \frac{1}{2}x^T K^{-1} x \right)
    \frac{\prod_{(m,n) \in \mathcal{T}} f_m(x_m) f_n(x_n)}{\prod_{n \in
    \mathcal{T}} f_n(x_n)^{d_n-1}}
\end{equation}
we will then approximate the latter part by a Gaussian restricted to the tree
\begin{equation}
    p(x) = \frac{1}{Z} \exp\left(- \frac{1}{2}x^T K^{-1} x \right)
    \frac{\prod_{(m,n) \in \mathcal{T}} g_{m,n}(x_m,x_n)}{\prod_{n \in
    \mathcal{T}} g_n(x_n)^{d_n-1}}
\end{equation}
The moment matching algorithm can then be developed requiring matching of
the moments $\langle x_m x_n \rangle \in \mathcal{T}$, in addition to the
single variable moments $\langle x_n \rangle$ and $\langle x_n^2 \rangle$. 

For discrete random variables, Qi and Minka have proposed a method treeEP
which is in a similar spirit (\citealp{Minka:TSA}). It involves a non Gaussian
approximation with consistency of the moments on both edges and vertices
required.

\subsection{Expansion methods}
The aim will be to demonstrate the connection between the approximate
distribution $q$ and the true distribution $p$ as an expansion in some small
terms. The expansion can indicate sufficient conditions for EP to succeed,
as well as be used in practice to improve estimation. In the selection of
expansion methods, we do not assume that the interactions reveal any inate
structure -- if this were so we might select or refine the approximation
tailored to account for this clustering.

Consider the difference between the exact and approximate probability
distributions for the standard approximation on marginal statistics, as
described by
\begin{equation}
    p(x) = \frac{1}{Z} \prod_n f_n(x)\;; \qquad  q(x) = \frac{1}{Z_q} \prod_n g_n(x)\;.
\end{equation}

A Gaussian interacting part can be included as $f_0 = g_0$. The tilted
distribution we recall as
\begin{equation}
    q_n(x) = \frac{1}{Z_n} \left(\frac{q(x) f_n(x)}{g_n(x)} \right)\;.\label{eq:tilted}
\end{equation}

Solving for $f_n$ yields
\begin{equation}
    \prod_n f_n(x) = \frac{Z_n q_n(x) g_n(x)}{q(x)} = Z_{EP} \, q(x) \prod_n
    \left(\frac{q_n(x)}{q(x)} \right)\;,
\end{equation}
where we introduce the definition of the EP partition function
\begin{equation}
    Z_{EP} = Z_q \prod_n Z_n\;.
\end{equation}

In terms of the function
\begin{equation}
    F(x) = \prod_n \left(\frac{q_n(x)}{q(x)} \right)
\end{equation}
we can write
\begin{equation}
    p(x) = \frac{1}{Z} \prod_n f_n(x) = \frac{Z_{EP}}{Z} q(x) F(x)
\end{equation}
the ratio of the true to approximate partition functions defines the
normalization constant
\begin{equation}
    \frac{Z}{Z_{EP}} = \int q(x) F(x) dx
\end{equation}
Note the term $F(x)$ should be close to $1$, and $Z\approx Z_{EP}$, when the
approximation method works well.

\subsubsection{An expansion in $q_n(x)/q(x) -1$}
So long as the approximation is meaningful, the tilted and fully
approximated distributions should be close, hence we can take the quantity
\begin{equation}
    \epsilon(x) = \frac{q_n(x)}{q(x)} - 1
\end{equation}
to be typically small. The exact probability is then
\begin{equation}
    p(x) = q(x) \frac{1 + \sum_n \epsilon_n(x) + \sum_{n_1<n_2}
        \epsilon_{n_1}(x) \epsilon_{n_2}(x) + \ldots}{1 + \sum_{n_1 < n_2}
        \langle \epsilon_{n_1}(x) \epsilon_{n_2}(x) \rangle_q + \ldots }
        \label{eq:expansion1}
\end{equation}
The ratio of the partition functions (the denominator) is expanded as
\begin{equation}
    R = 1 + \sum_{n_1 < n_2} \langle \epsilon_{n_1}(x) \epsilon_{n_2}(x)
        \rangle_q + \sum_{n_1<n_2<n_3 } \langle
        \epsilon_{n_1}(x)\epsilon_{n_2}(x)\epsilon_{n_3}(x)\rangle \;.
\end{equation}
At first order $\sum_n \langle \epsilon_n(x)\rangle_q = 0$, by
normalization.  Thus, the first order correction to the probability in
$\epsilon_n$ is particularly simple in that it {\em does not require the
computation of expectations}
\begin{equation}
    p(x) \approx \sum_n q_n(x) - (N-1)q(x) \;.
\end{equation}

A related calculation by Cseke and Heskes leads to a similar correction
identity for marginal probabilites (\citealp{Cseke:AMLGM})
\begin{equation}
    p(x_i) \approx q_i(x_i) \prod_{j (\neq i)} \int d x_j q(x_j|x_i) \frac{f_j(x)}{g_j(x)}\;.
    \label{eq:correc_marginal}  
\end{equation}

For an illustration of this expansion we consider a class of models called
Bayesian mixture of Gaussians which is used to fit a mixture of $K$
Gaussians to data points $\zeta_n$.  The latent variables of the model are
$x = \{\pi_\kappa,\mu_\kappa,\Gamma_\kappa\}_{\kappa=1}^K$ which gives the
weight, the mean and the covariance matrix of the Gaussians in the mixture.
Given a set of data $\zeta_n$ $n=1,\ldots, N$, and a prior, we will be
interested in inferring these parameters. The likelihood given a data point
$\zeta_n$ we will describe as \begin{equation}
    f_n(x) = \sum_\kappa \pi_\kappa \mathcal{N}(\zeta_n; \mu_\kappa,\Gamma_\kappa^{-1}) \;,
\end{equation}
where $\mathcal{N}(a;b,c)$ denotes a Gaussian distribution of the random
variable $a$ with mean $b$ and covariance $c$. A convenient choice for the
prior is a product of Dirichlet and Wishart distributions
\begin{equation}
    f_0 = \mathcal{D}(\pi) \prod_k \mathcal{W}(\mu_k,\Gamma_k)
\end{equation}

From the product of the prior, and likelihoods for different data, we obtain
the posterior
\begin{equation}
    p(x | \zeta_1,\ldots,\zeta_N) = \frac{1}{Z}\prod_{n} f_n(x)\;.
\end{equation}

The approximation we make assumes the same form as the prior
\begin{equation}
    q(x) = \mathcal{D}(\pi) \prod_k \mathcal{W}(\mu_k,\Gamma_k)\;.
\end{equation}

A second class of models is given by the so called Gaussian process popular
in the area of machine learning (\citealp{Rasmussen:GPML}). Here one assumes
latent variables with a joint prior distribution given by a Gaussian.

The Gaussian process classification problem involves a likelihood which
leads to a non-Gaussian posterior probability $p(x)$. The classification can
however be modelled by a Gaussian yielding the approximation $q(x)$; thus
the structure of the problem is suitable for EP.

\begin{figure}[h!tbp]
    \begin{center}
    \includegraphics[width=0.8\linewidth]{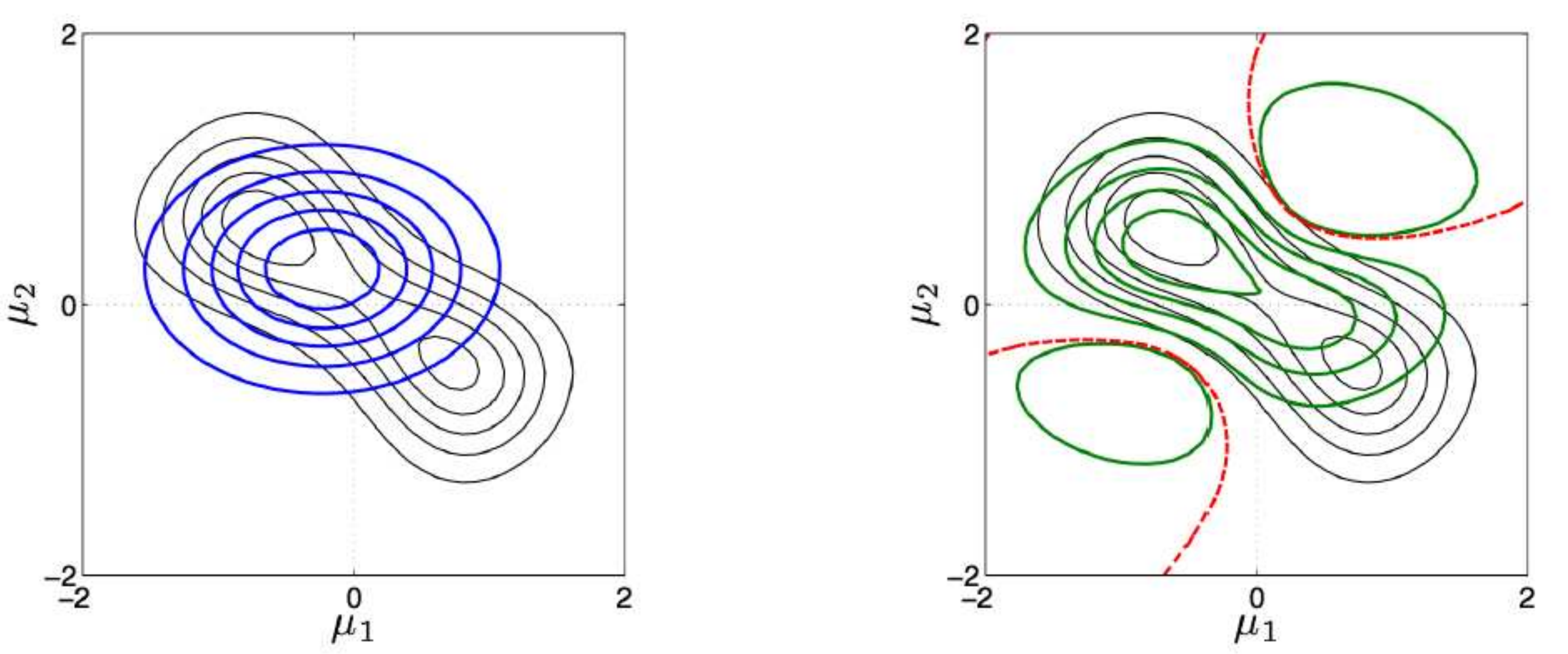}
    \caption{\label{fig:11} We generate a toy mixture model of two Gaussians,
        where only the means $\mu_{1,2}$ of the Gaussians are assumed to be unknown
        latent variables. Posterior probability distribution contour plots are
        presented (grey) alongside the approximation. Left (blue) is the EP
        estimate, right (green) is the leading order correction. Red indicates a
        region in which the corrected probability becomes unphysical (negative).  }
    \end{center}
\end{figure}

\begin{figure}[h!tbp]
    \begin{center}
    \includegraphics[width=0.4\linewidth]{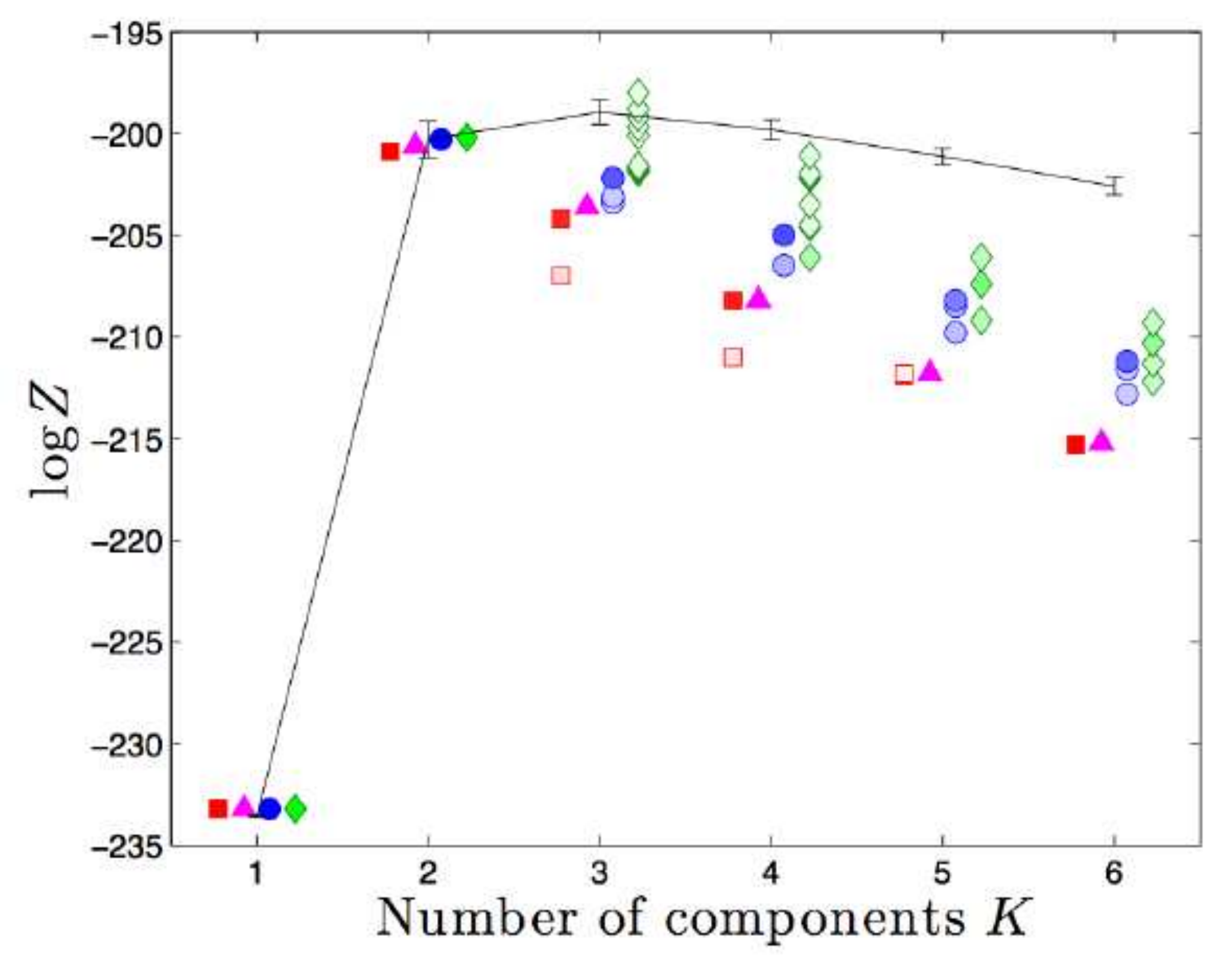}
    \caption{\label{fig:12} Approximations to a K-component Gaussian mixture
        model parameterized by a standard data set (the acidity data set) which
        gives data on acidity levels across 155 US lakes. Black with error bars
        indicates the Monte-Carlo estimate to the log partition function.
        Significantly faster approximations are: Variational Bayes (red
        squares) (\citealp{Attias:VBF}), Minka's $\alpha=\frac{1}{2}$-divergence method
        (magenta triangles)~(\citealp{Minka:DM}), standard EP(blue circles) and EP with
        the $2^{nd}$ order corrections (\ref{eq:expansion1}). Symbol intensity
        indicates the frequency of the outcome based on 20 independent runs. Figure
        extracted from~\citealp{Opper:IEP}.}
    \end{center}
\end{figure}

\begin{figure}[h!tbp]
    \begin{center}
    \includegraphics[width=0.4\linewidth]{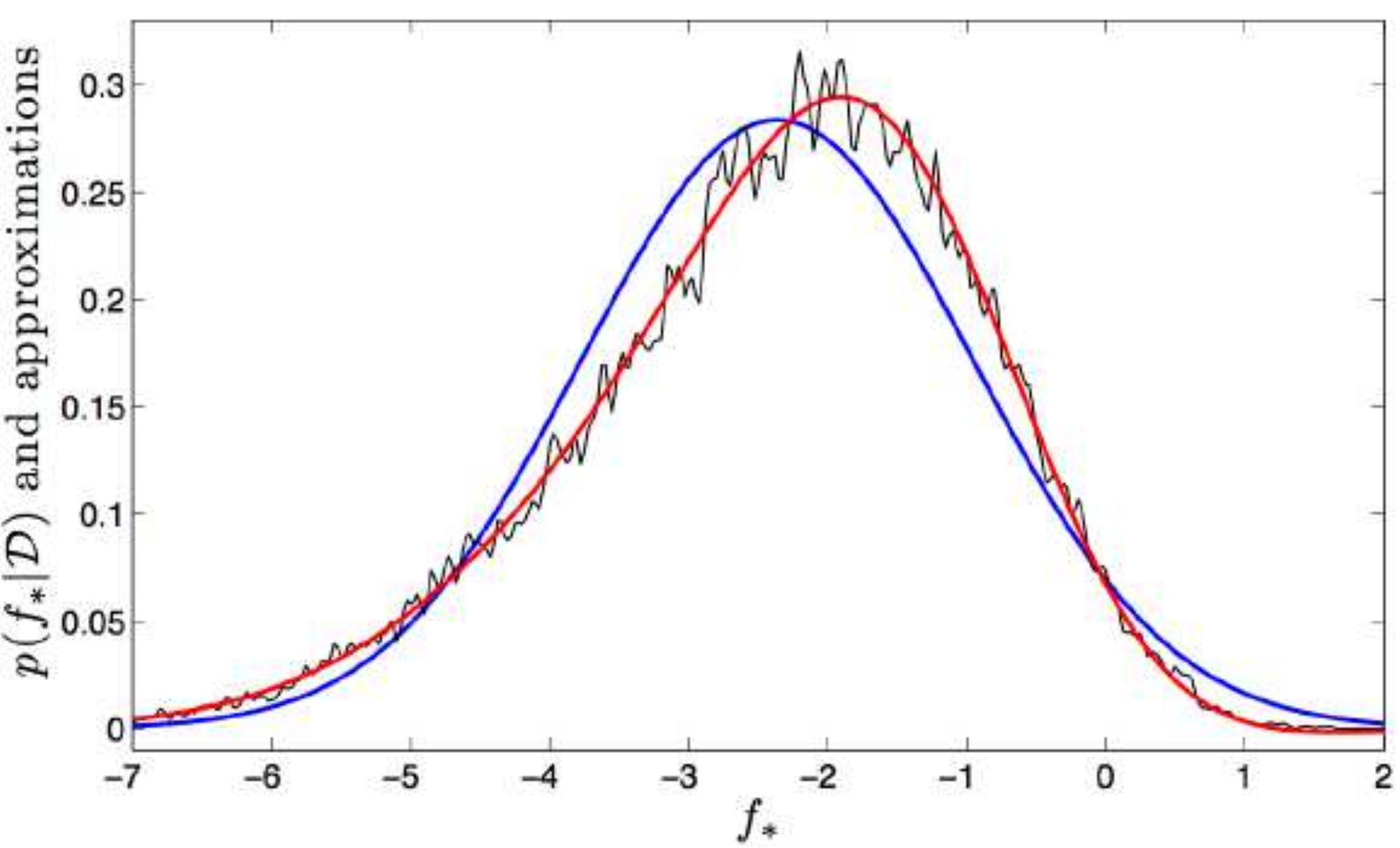}
    \caption{\label{fig:13} Marginal posterior for a toy Gaussian process
        classification. A MCMC approximation (grey) compared to faster
        approximations: standard EP (blue), and EP with the first order correction
        (red) (\ref{eq:expansion1}).}
    \end{center}
\end{figure}

Within the context of mixture of Gaussians and Gaussian process
classification the expansion method (\ref{eq:expansion1}) can be very
effective, as shown in figures \ref{fig:11}-\ref{fig:13}.

\subsubsection{An expansion in cumulants}
The cumulants of a distribution are an alternative description, they
represent a convenient framework for expansions about Gaussians.

Consider the latent Gaussian model, where
\begin{equation}
    q(x) \propto \exp\left[-\frac{1}{2}x^T K^{-1} x \right]\prod_n
        \exp(\gamma_n x_n - \lambda_n x_n^2/2) \label{eq:Gaussianlatent}
\end{equation}
is used to approximate 
\begin{equation}
    p(x) = \frac{1}{Z} \exp\left[-\frac{1}{2}x^T K^{-1} x \right]\prod_n
    f_n(x_n) 
    \label{eq:intractable model}
\end{equation}

The tilted distribution (\ref{eq:tilted}) is required to match $q(x)$ in the
first and second cumulant.  The error can be quantified by the ratio of the
parition functions ($R=Z/Z_{EP}$), which can be expressed
\begin{eqnarray}
    R &=& \int q(x) \prod_n\left(\frac{q_n(x)}{q(x)} \right) \\ 
    &=& \int q(x) \prod_n\left(\frac{q(x_{\setminus n}|x_n)
        q_n(x_n)}{q(x_{\setminus n}|x_n)  q(x_n)} \right) \\ 
    &=& \int dx q(x) \prod_n \left(\frac{q_n(x_n)}{q(x_n)} \right)
\end{eqnarray}
where notation $x_{\setminus n}$ is the set of all variables excluding $n$,
and so $q(x_{\setminus n} | x_n)$ is the conditional probability. So long as
$q_n$ is Gaussian the ratio is one, the errors can be accounted for by the
higher order cumulants of the tilted distribution, so these are good
candidates for an expansion.

In terms of the Fourier transform $\chi_n(k)$ of the marginal tilted
distribution, the original distribution is described by
\begin{equation}
    q_n(x_n) = \int_{-\infty}^{\infty} \frac{d k}{2 \pi} \exp(- i k x_n) \chi_n(k)\;,
\end{equation}

In the Fourier basis we have a simple expression for all the neglected
cumulants $r_n(k)$ (third order and higher) as
\begin{equation}
    \log \chi_n(k) = \sum_{l}(i)^{l} \frac{c_{nl}}{l!} k^l = i m_n k - S_n
    \frac{k^2}{2} + r_n(k)\;.
\end{equation}

The first two moments are equal in the tilted and untilted distribution by
the moment matching conditions. The ratio of the two distributions can be
found by resubstitution, in terms of an integral
\begin{equation}
    \frac{q_n(x_n)}{q(x_n)} = \sqrt{\frac{S_{nn}}{2
    \pi}}\exp\left(\frac{(x_n-m_n)^2}{2
    S_{nn}}\right)\int_{-\infty}^{\infty} \frac{d k}{2 \pi} \exp(- i k x_n)
    \chi_{n}(k) \;.
\end{equation}

Using the linear change of integration variable
\begin{equation}
    \nu_n = k + i \frac{x_n - m_n}{S_{nn}}\;,
\end{equation}
we can abbreviate
\begin{equation}
    \frac{q_n(x_n)}{q(x_n)} = \int_{-\infty}^{\infty}d \nu_n
    \sqrt{\frac{S_{nn}}{2 \pi}} \exp\left[-\sum_n \frac{S_{nn}\nu_n^2}{2}
    \right] \exp\left[r_n \left(\nu_n - i \frac{x_n - m_n}{S_{nn}}\right)
    \right]\;.
\end{equation}

The ratio of partition functions
\begin{equation}
    R = E_{q}\left[\prod_n \left(\frac{q_n(x_n)}{q(x_n)}\right)\right]
\end{equation}
requires an integral over the multivariate Gaussian $q(x) = \mathcal{N}(x;
m, S)$, in addition to $\nu_n$. We have a double integration over an
expression that depends on a weighted sums of the parameters $\nu_n$ and
$x_n$. Since these are Gaussian variables, and the sum of two Gaussian
variables is itself Gaussian, we can replace the weighted sums by a new
complex Gaussian random variable
\begin{equation}
    z_n = \nu_n - i \frac{x_n - m_n}{S_{nn}}
\end{equation}
of zero mean, and with a distribution described by covariances
\begin{equation}
  \langle z_n^2 \rangle_z = 0\;;\qquad
  \langle z_m z_n \rangle_z = -\frac{S_{mn}}{S_{nn}S_{mm}}\;,
\end{equation}

The double integral is thereby replaced by a single integral
\begin{equation}
  \frac{Z}{Z_{EP}} = \left\langle\exp\left[\sum_n r_n(z_n) \right]\right\rangle_z\;.
\end{equation}
Assuming that the cumulants $c_{ln}>2$ are small we can make a power series
expansion in $r_n(z_n)$, to obtain 
\begin{eqnarray}
    \log(R) &=& \frac{1}{2} \sum_{m \neq n} \langle r_m r_n \rangle_z + \ldots \\
    &=& \sum_{m \neq n} \sum_{l \geq 3}
        \frac{c_{ln}c_{lm}}{l!}\left(\frac{S_{mn}}{S_{nn}S_{mm}} \right)^l +
        \ldots 
    \label{eq:expansion2}
\end{eqnarray}
The self-interaction (diagonal) term is absent, which indicates that
corrections may not scale with $N$, a desirable property atleast in so far
as theory is concerned.

\begin{figure}[h!tbp]
    \begin{center}
    \includegraphics[width=0.4\linewidth]{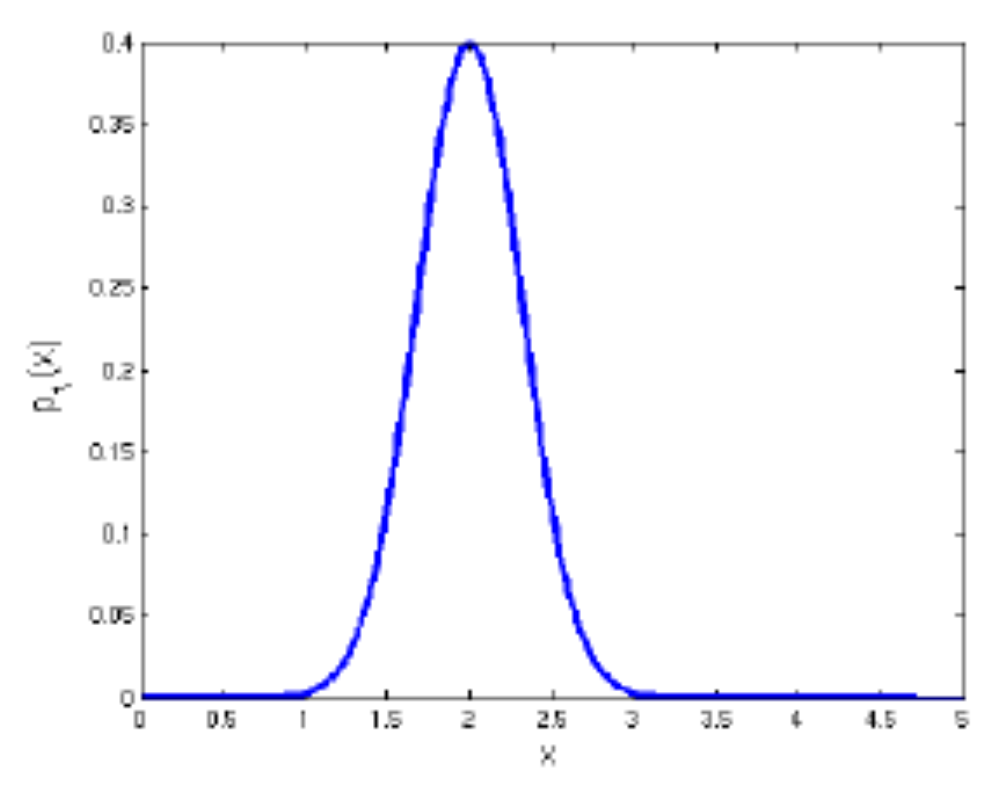}
    \caption{\label{fig:20} A toy problem considering classification
        likelihoods $f_i(x_i)= \Theta(y_i x_i)$. Provided posterior variance is
        small compared to the mean, the portion of the Gaussian extending across the
        threshold at zero, is small. The truncation of the Gaussian in the tail
        results in only a small modification of the cumulants in the tilted function
        $q_i(x_i)$.}
    \end{center}
\end{figure}

The cumulant expansion allows us to consider the scenarios in which EP may
be accurate. From the calculation we note that the correction is small if
either the cumulants $c_{ln}$ are small, which is often true in
classification problems (see for example figure \ref{fig:20}), or the
posterior covariances $S_{mn}$ are small for $m\neq n$. 

\begin{figure}
    \begin{center}
    \includegraphics[width=0.8\linewidth]{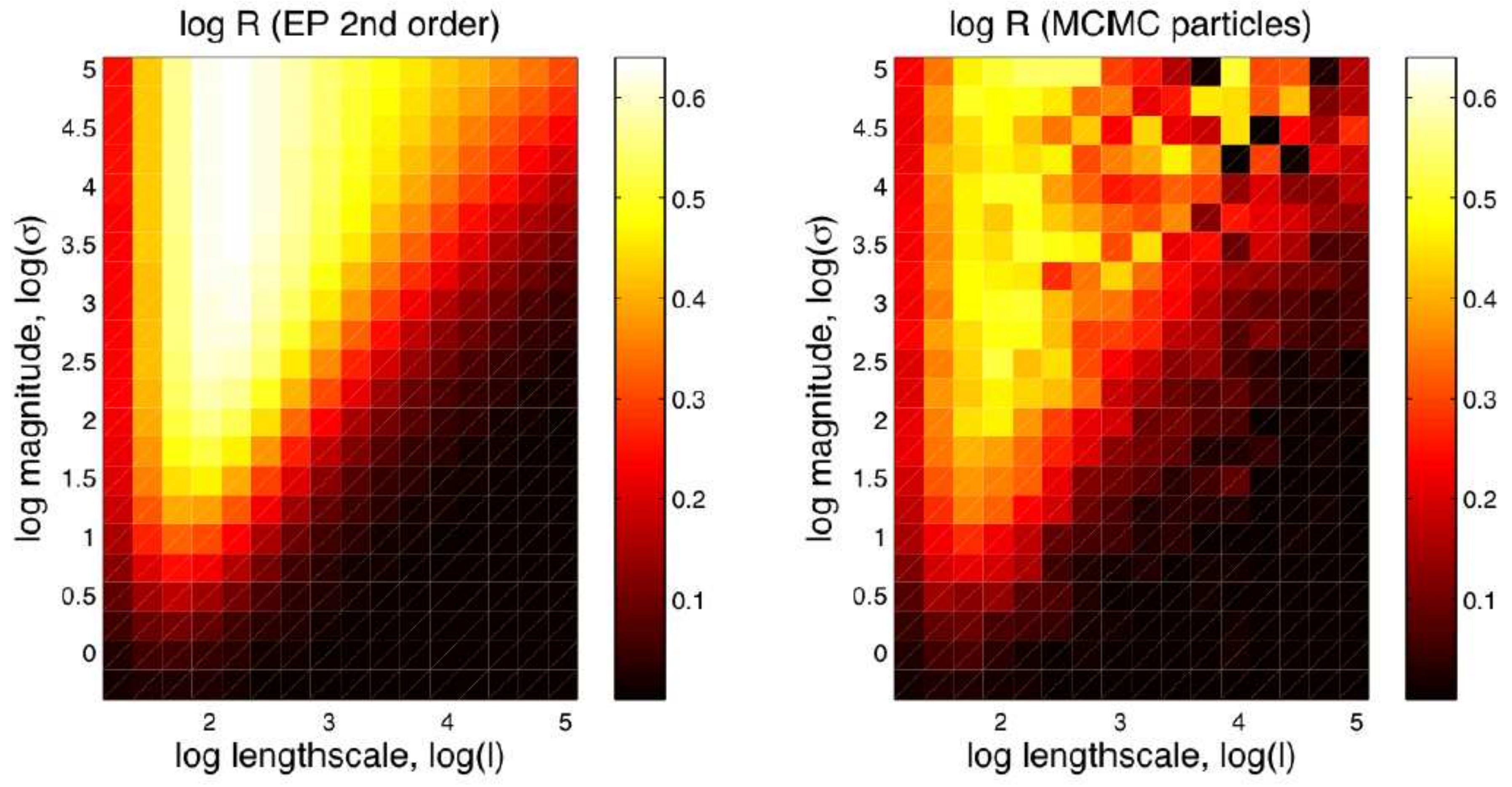}
    \caption{\label{fig:21} Analysis of Gaussian Process classification on
        US postal service data. The data set consists of 16 x 16 grayscale
        images (real valued vectors with components on $[0,1]$) of handwritten
        digits ($[0,9]$). The free energy is already well approximated by $EP$. The
        remaining difference in the free energy ($\log R$) is mostly accounted for
        by including $l \leq 4$ corrections (left), by comparison with the exact
        Monte Carlo evaluation (right). The exact evaluation is slower by orders of
        magnitude. The free parameters are a length scale $l$, and the width of the
        Gaussian prior $\sigma$. Figure extracted from~\citealp{Opper:PC}, see
        also~\citealp{Rasmussen:GPML}.}
    \end{center}
\end{figure}

\begin{figure}
    \begin{center}
    \includegraphics[width=0.4\linewidth]{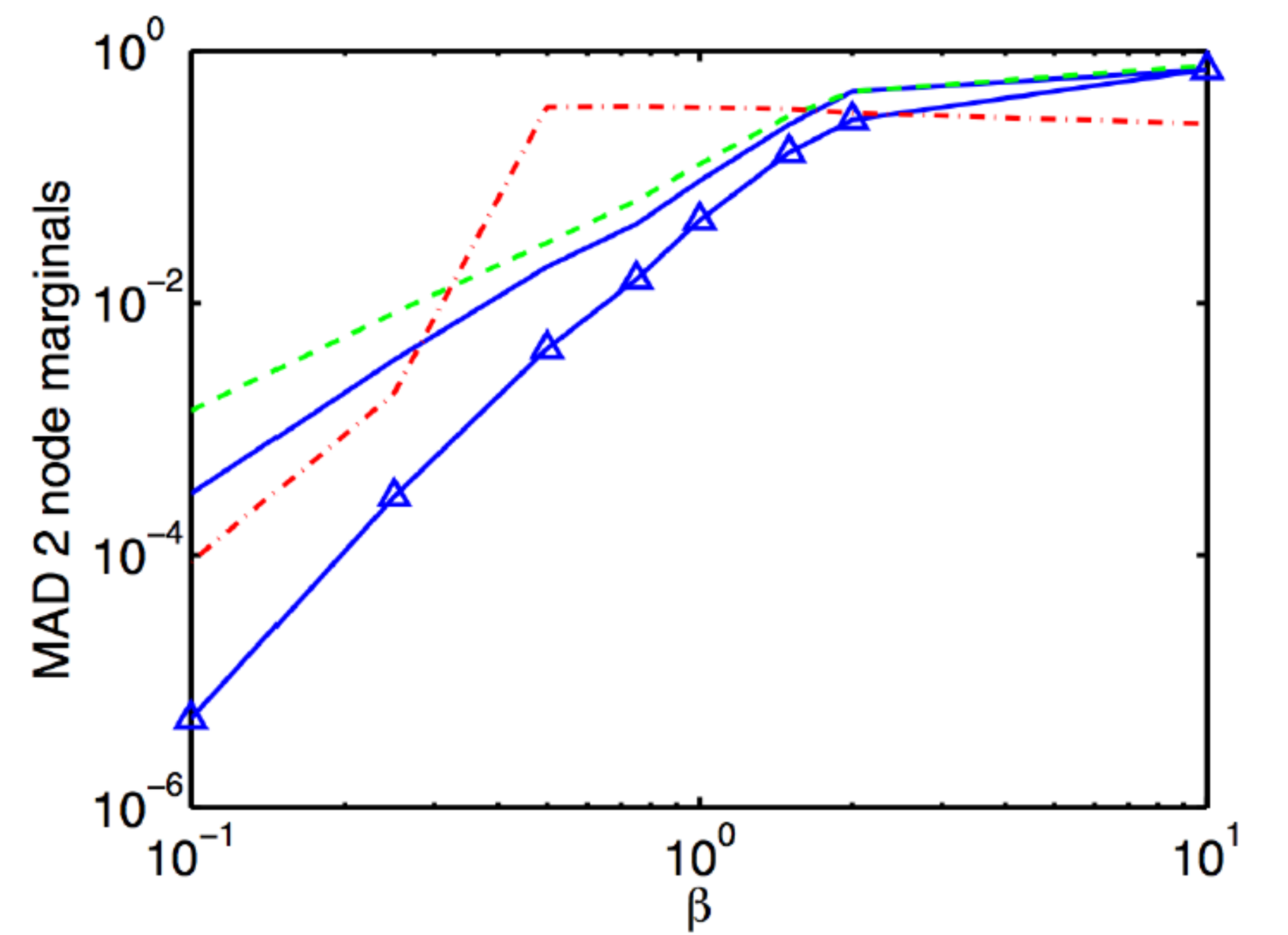}
    \includegraphics[width=0.4\linewidth]{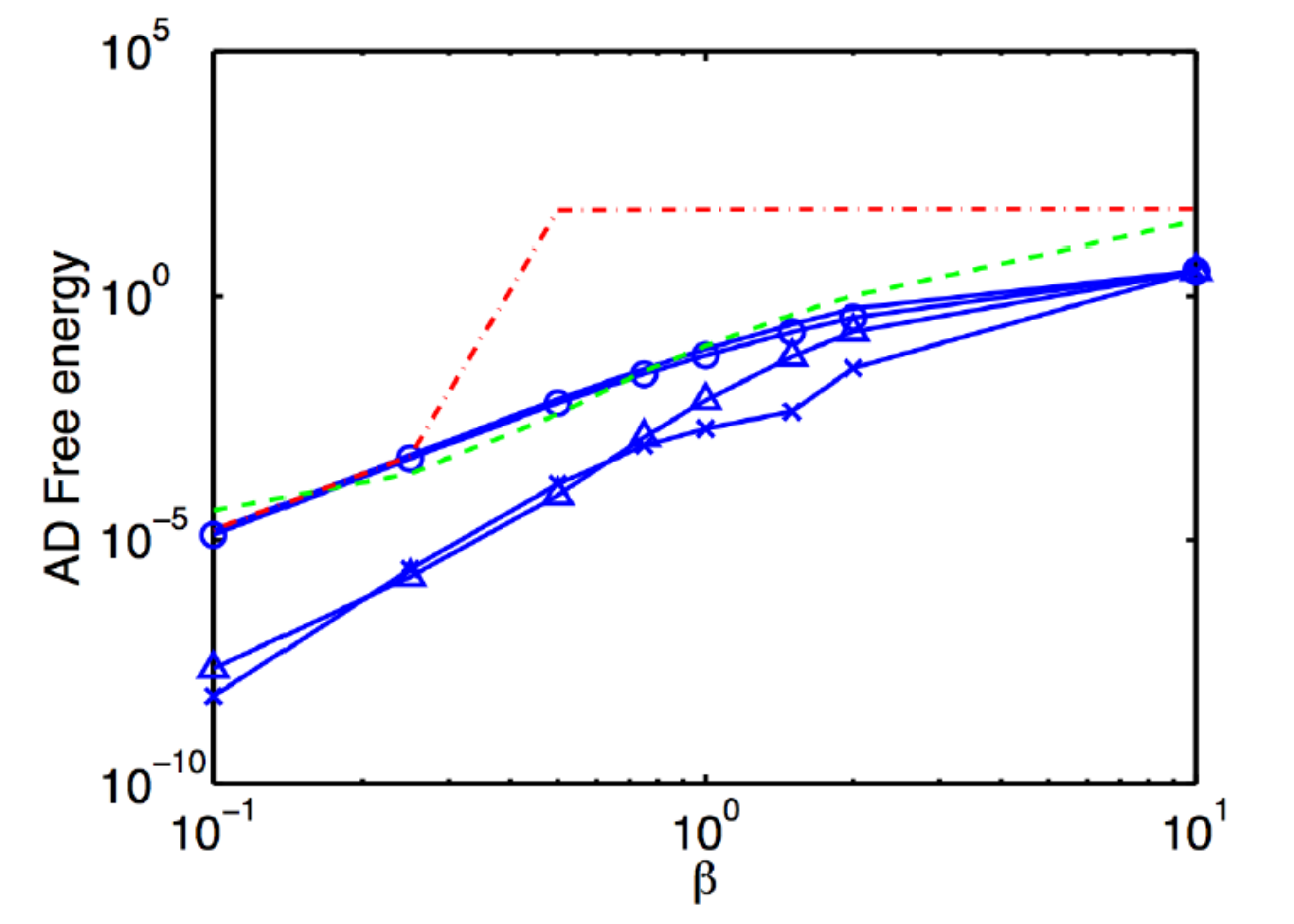}
    \caption{\label{fig:22} An Ising spin model of $N$ spins, with zero
        field and random couplings of variance $\beta^2$. The absolute
        difference (AD) between exactly calculated values and approximation values
        for respectively the 2 node marginals (left) and true free energy ($-\log
        Z$) (right). Approximate methods: Loopy belief propagation (LBP) (dashed
        green), Generalized LBP [with shortest loops included in outer region]
        (dashed red), standard EP (blue no symbol). The second order cumulant
        corrections (\ref{eq:expansion2}) with $l\leq$ 3 (circle), 4 (triangle), 5
        (cross) ameliorate the error. Figure extracted from ~\citealp{Opper:IEP}.}
    \end{center}
\end{figure}

\begin{figure}
    \begin{center}
    \includegraphics[width=0.6\linewidth]{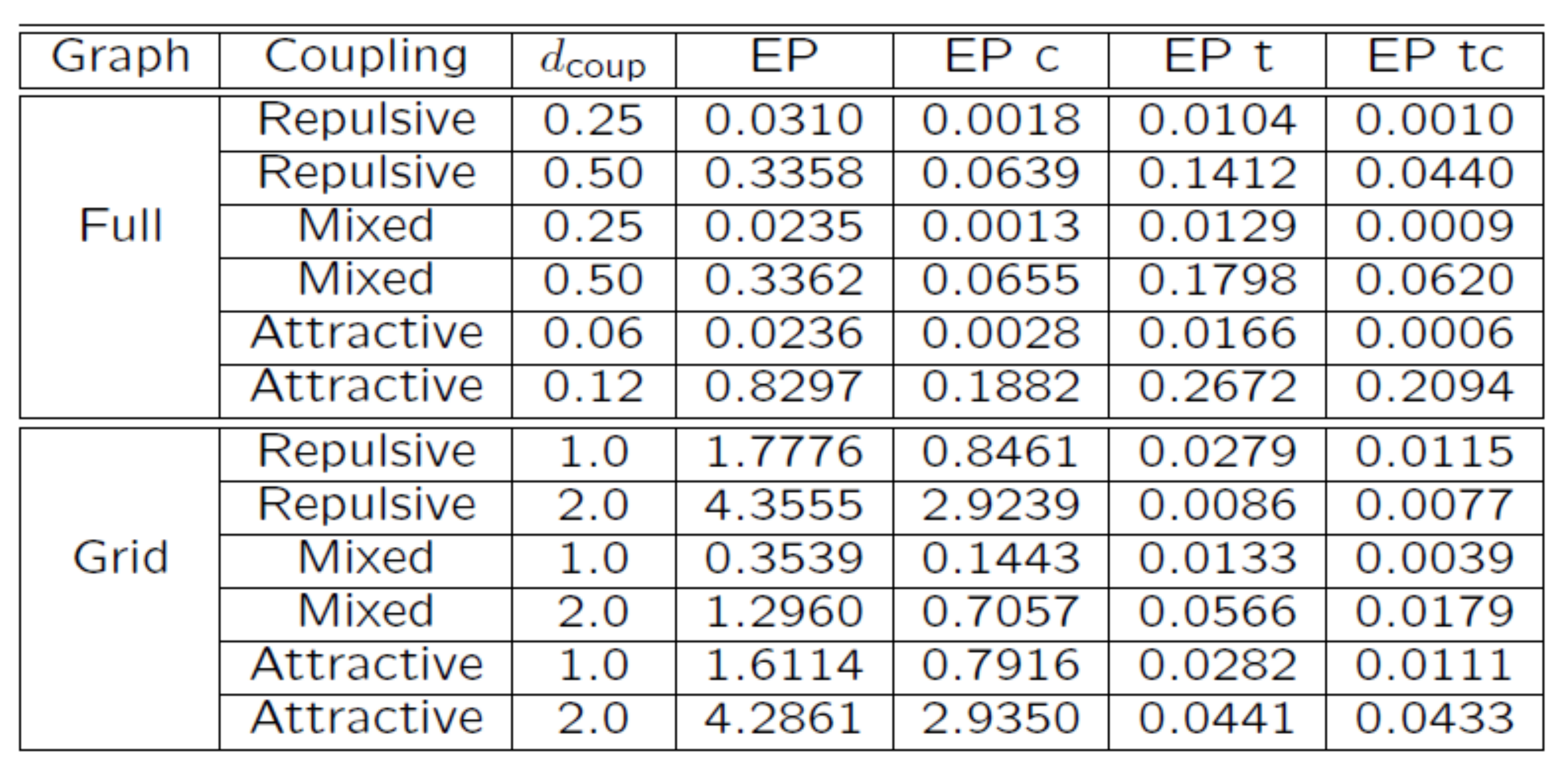}
    \caption{\label{fig:24} 
        In the Wainwright-Jordan set-up there are $N = 16$ Ising spins which are
        either (upper table) fully connected or (lower table) connected to nearest
        neighbors in a $4$ by $4$ grid. Fields are sampled uniformly at random in
        $[-0.25,0.25]$ and couplings are sampled from distributions of width
        $[a-d_{coup},a+d_{coup}]$ where $a=d_{coup}$ is the Attractive case, $a=0$
        the Mixed case and $a=-d_{coup}$ the repulsive case. Average absolute
        deviation of $\log Z$ are shown, comparing standard EP (EP), EP with tree
        corrections (EP t), and the same approximations with $l=4$ second order
        cumulant corrections (EP c and EP tc respectively). Table extracted from
        ~\citealp{Opper:IEP}.}
    \end{center}
\end{figure}

\begin{figure}
    \begin{center}
    \includegraphics[width=0.4\linewidth]{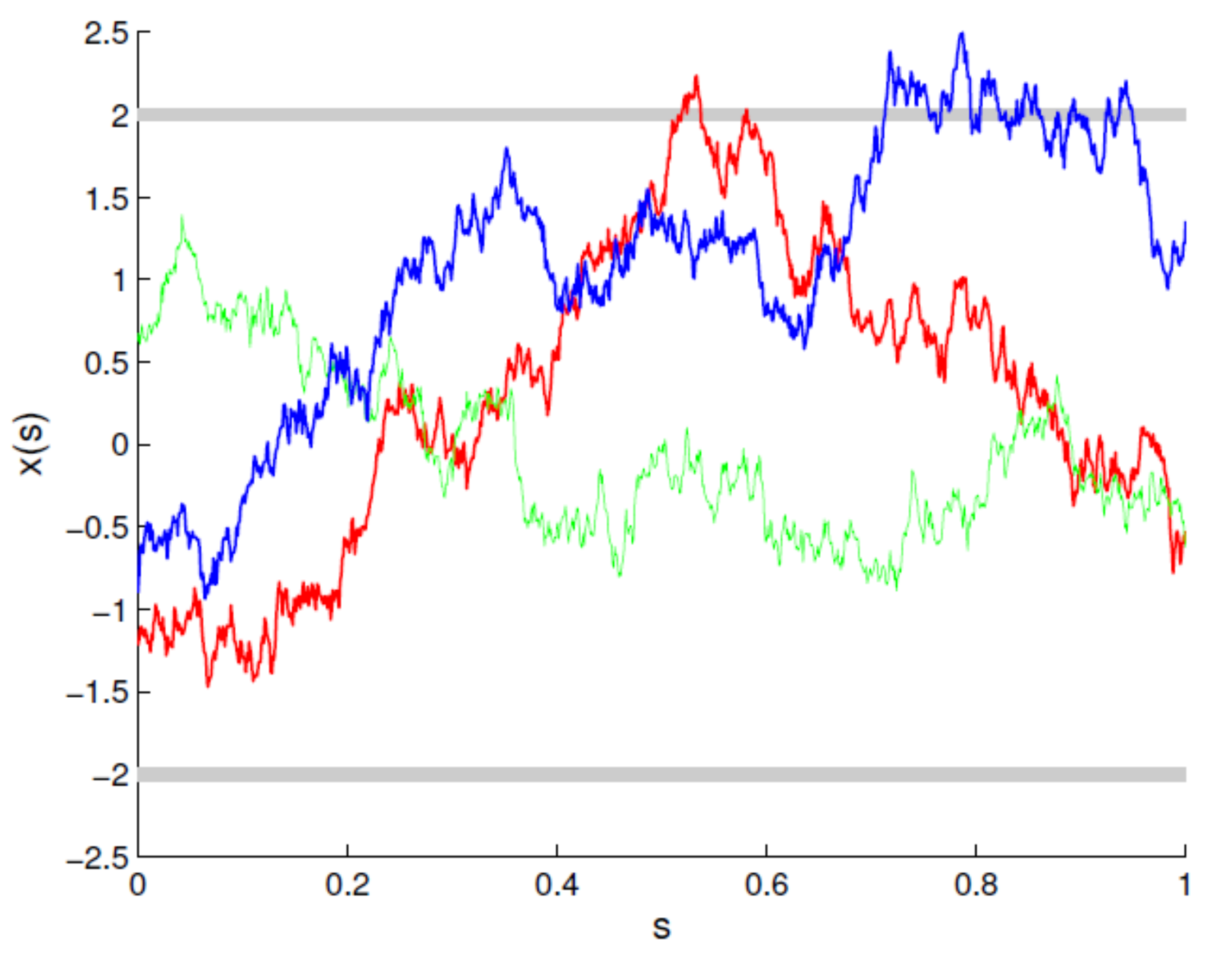}
    \includegraphics[width=0.4\linewidth]{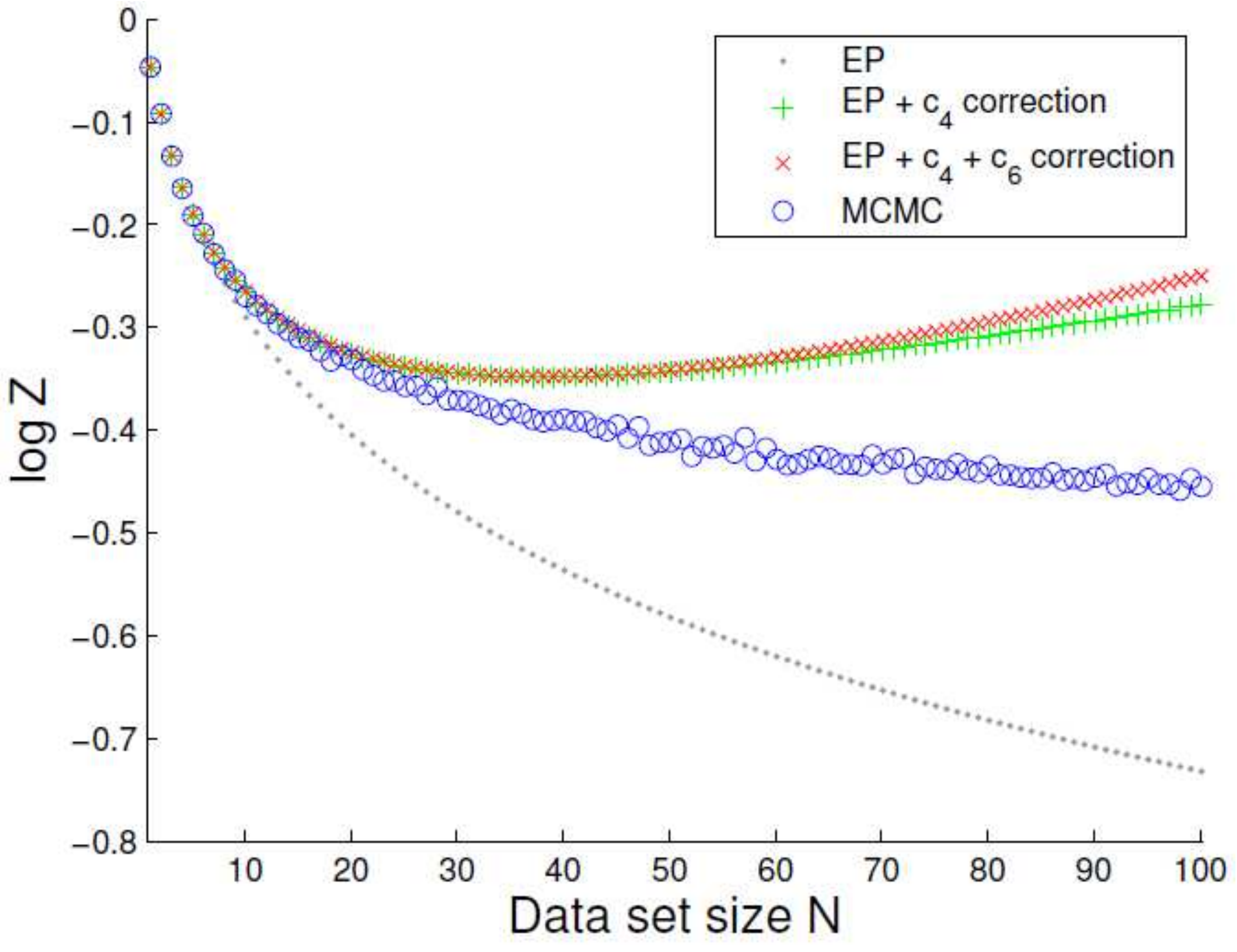}
    \caption{\label{fig:25} Gaussian process in a box: We are interested in
        the probability distribution over a Gaussian process constrained to
        remain in the interval $[-a,a]$ at all times: $p(x) = \frac{1}{Z}
        \prod_n \theta(a-|x_n|)\mathcal{N}(x;0,K) $, $\theta$ is the
        Heaviside function. Realizations of $x$ are shown left (of which one
        is valid).  $Z$ is the fraction of processes that remain in the box,
        as the number of points $n$ increases we anticipate convergence.  EP
        strongly underestimates the number of solutions, whilst the corrections,
        after initial improvements, significantly overestimate.}
    \end{center}
\end{figure}

Applications of the cumulant expansions are shown in figures
\ref{fig:21}-\ref{fig:25}. In the first four cases we see significant
improvements in the estimation. In the final example of a Gaussian process
in a box the EP estimate becomes increasingly inaccurate as the number of
observations $N$ increases, the corrections about this result do not much
improve the estimation; standard EP is clearly not universally useful, a
better approximation to capture the problem structure is required before
expansion methods can be useful.

\section{Some advanced topics and applications}
\begin{figure}
    \begin{center}
    \includegraphics[width=0.4\linewidth]{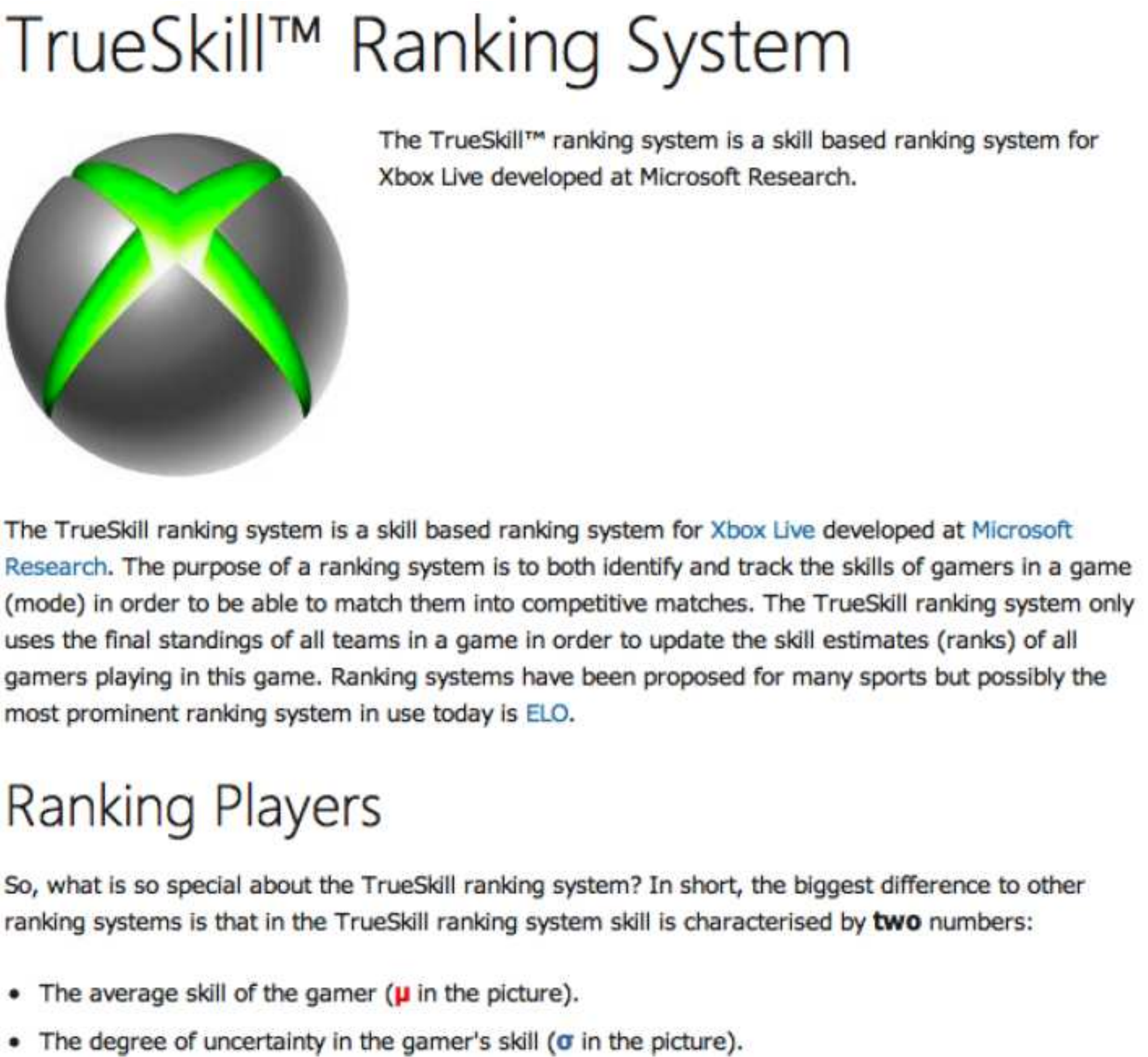}
    \caption{\label{fig:27} Trueskill is a practical application of the EP
    algorithm, amongst others being pursued by Minka et al. at Microsoft
    research:
    http://research.microsoft.com/en-us/um/people/minka/papers/ep/roadmap.html}
    \end{center}
\end{figure}

It is worth noting that already in a form closely related to that presented
there are many applications of EP, examples being Trueskill and recommender
systems at Microsoft (see figure \ref{fig:27}). In the Trueskill application
one has a set of data which is the outcome of games between players matched
online. One can imagine that each player has a certain positive skill level
(that might be time dependent) modeled by a latent variable $x_i$ (the true
skill). The discrete (binary) outcome of a game is the data, victory or
defeat might be modeled by the sign of $x_i-x_j+\nu$, where $\nu$ is some
noise. This defines a simple likelihood suitable for the methods outlined in
previous sections, and EP is being used in practice to determine player
skill levels. The true model is more complicated, since it involves a
dynamic element, but it is still simple enough to process an estimated
$O(10^6)$ games for $O(10^5)$ players every day in real time.

In the remainder of this section we outline some other advanced applications
and generalizations of the methods presented in earlier sections. These
examples show how EP can be used out of the box in combination with other
analytical techniques such as the replica method, and for continuous time
and other frameworks outside the standard machine learning context.

\subsection{Bootstrap estimation for Gaussian regression models, an
application of the replica method}

A standard problem in inference is to determine the mean square error of a
statistical estimator ${\hat E}[x_i | D]$, such as a method for predicting
the mean or variance of a distribution. We now outline a method to combine
EP with the replica method to reduce the complexity of bootstrap
estimation (\citealp{Malzahn:Regression}). 

In practice we do not have access to the distribution, only $N$ data points
$D_0$; to get around this problem {\em we pull ourselves up by our
bootstraps} -- using a subset of the data for training, whilst the remaining
set constructs an empirical distribution $D$. The {\em pseudo data} that
builds the empirical distribution is determined via resampling with
replacement.

Consider a vector $m_i$ to represent the sampling process: each data point
can be included $0$, $1$ or several times in forming the empirical
distribution $D$.  Those elements not occupied $m_i=0$ are independent of
the empirical distribution, and can be used as the test set. The test error
(the Efrons estimator) can then be constructed as
\begin{equation}
    \epsilon(m) = \frac{1}{N}\sum_{i}
    \frac{E_D\left[\delta_{m_i,0}\left(E[x_i|D] - y_i\right)^2
    \right]}{E_D\left[\delta_{m_i,0}\right]}
    \label{eq:epsm}
\end{equation}

One problem with the scheme is that for each sample we must reevaluate the
error, and many samples are required for a robust estimate.

One can consider an approximation to this quantity using statistical physics
insight as discussed in other courses.  Asymptotically occupation numbers
$m_i$ become independently and identically distributed Poisson random
variables of mean $m/N$, so the denominator in (\ref{eq:epsm}) is
$\exp(-m/N)$. Still for a particular sample we can write
\begin{equation}
    \epsilon_n(m) = \frac{1}{\exp(-m/N) N} \sum_{i=1}^N
    E_D\left[\delta_{m_i,0}Z^{n-2}\int \prod_{j=1,2} [dx^{(j)}
    p_0(x^{(j)})P(D|x^{(j)}) (x_i^{(j)} - y_i)]\right]
    \label{eq:epsm_avm}
\end{equation}

The idea of the replica method is to compute $\epsilon_n(m)$ (approximately)
analytically for integer $n>2$ and then to take the limit $n\rightarrow 0$.
$Z$ is the partition function for a single replica, the full calculation is
found in \citealp{Malzahn:Regression}. We approximate with EP and take the
limit $n\rightarrow 0$ to approximate the Efrons estimator (\ref{eq:epsm}).
The analytical approximation to $E_D$ saves significantly on the
computation.
\begin{figure}
    \begin{center}
    \includegraphics[width=0.4\linewidth]{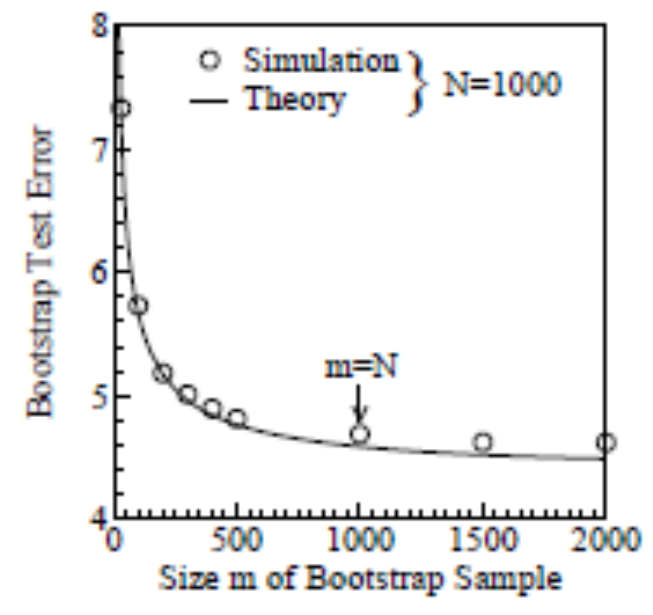}
    \includegraphics[width=0.4\linewidth]{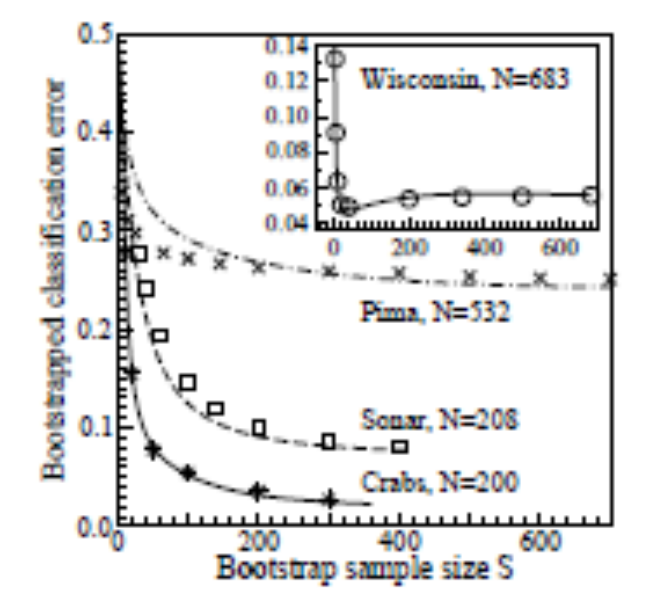}
    \caption{\label{fig:30} Example of results for Regression and Support
        Vector Machine Classification. Left: Averaged bootstrapped
        generalization error on the Abalone data (simulation: symbols, theory: lines; 
        each using the same $D_0$ of size $N=1000$). The Abalone data set relates
        a set of physical characteristics of an Abalones to their age, the later
        being the object of classified. Right: Average bootstrapped generalization
        error for hard margin support vector classification on different standard
        data sets (simulation: symbols, theory: lines). Figures are extracted
        respectively from ~\citealp{Malzahn:Regression,Malzahn:SVM}.}
    \end{center}
\end{figure}

\subsection{Gaussian approximations for generalized models}
For Gaussian latent models we have shown the power of the methods outlined.
We have however focused on pairwise models, we now identify one significant
generalization~(\citealp{opper_adaptive_2001}). We can generalize to a wider
class of models defined
\begin{equation}
    p(x) \propto \prod_{i=1}^N f_i(x_i) \exp\left[\sum_{i<j}^N x_i J_{ij} x_j
    \right] \prod_{k=1}^m F\left(\sum_i {\hat J}_{ik}x_k \right)
\end{equation}
Thus we are able to extend straightforwarly our consideration of latent
Gaussian models to classifiers as one example (e.g. a perceptron classifier
can take a form $F_k(x) = \Theta(\sum_i {\hat J}_{ik}x_k - b_k)$) and other
interesting models. Defining augmented random variables $\sigma=(x,{\hat
x})$ we can cast this model in the form
\begin{equation}
    p(\sigma) \propto \prod_i \rho_i(\sigma) \exp\left[\sum_{i<j}\sigma_i A_{ij}
    \sigma_j \right] \;.
\end{equation}
where the augmented coupling matrix is
\begin{equation}
    A = \left(\begin{array}{cc} J & {\hat J} \\ {\hat J}^T & 0 \end{array} \right)
\end{equation}
and $\rho_i(\sigma_i) = f_i(x_i) {\hat f}_i({\hat x}_i)$, where 
\begin{equation}
    f_i({\hat x}_i) = \int \frac{d h}{2 \pi i} \exp\left(- {\hat x} h\right) F_{i}(h)
\end{equation}

\subsection{Inference in continuous time stochastic dynamics}

To apply EP to continuous processes introduces new challenges, but these can
be overcome, we give one example~(\citealp{Cseke:AILDP}).  Suppose a prior
process (Ornstein-Uhlenbeck), variables $x$ have a continuous time index $t$
rather than the discrete $n$
\begin{equation}
    d x_t = (A_t x_t + c_t) dt + B_t^{1/2} dW_t\;.
\end{equation}
We can build the likelihood given {\em continuous} and discrete time
observations $y$.  The simplest case might be one in which we measure spike
counts, statistical log-Cox process would be described ($T_d$ are the set of
spike observation times) by the likelihood
\begin{equation} 
    p(\{y_{t_i}^d\},\{y_t^c\} | \{x_t\}) \propto \prod_{t_i \in T_d}
    p(y_{t_i}^d|x_{t_i}) \times \exp\lbrace-\int_{0}^{1} dt V(t,y_t^c,x^t)
    \rbrace \;.
\end{equation}
We wish to estimate the rate process.

\begin{figure}
    \begin{center}
    \includegraphics[width=0.8\linewidth]{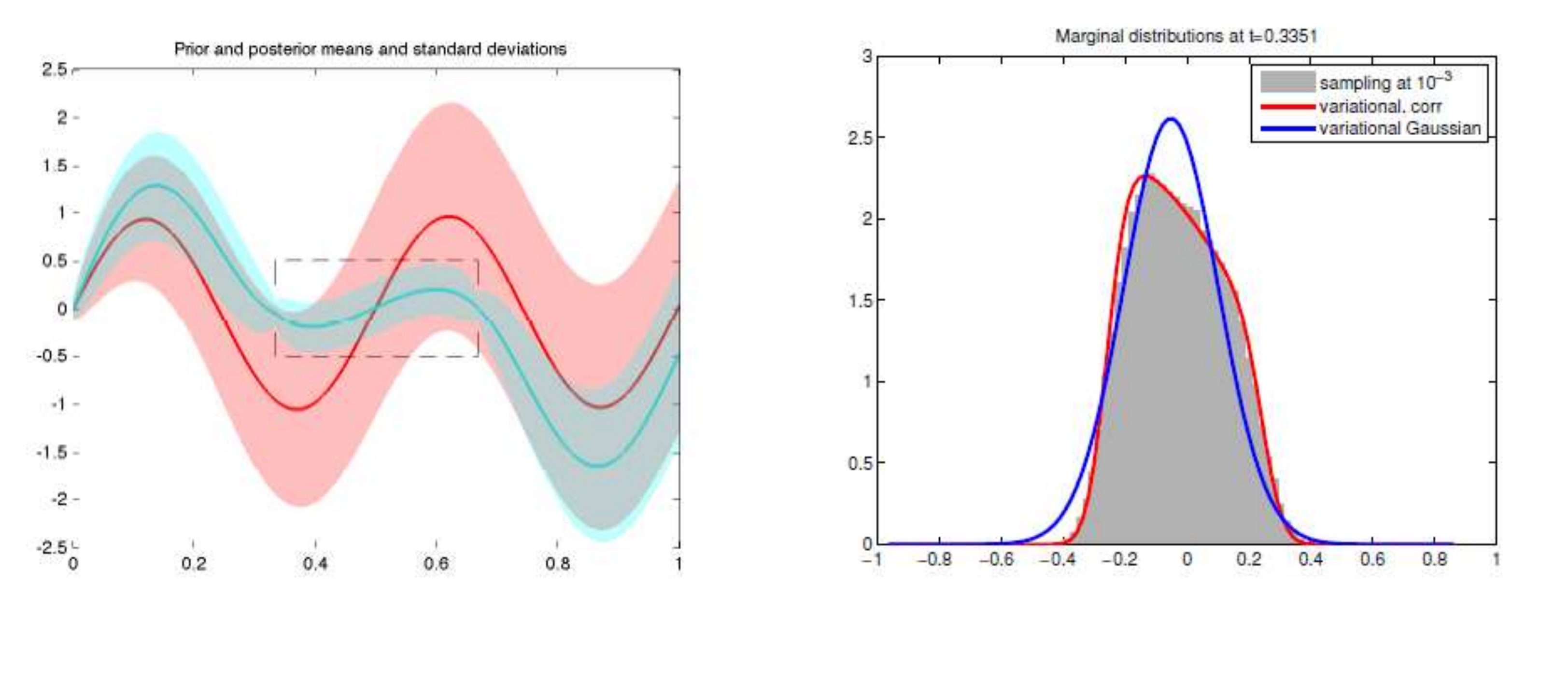}
    \caption{\label{fig:33} Left: A Gaussian prior with a periodic drift
        (orange, mean plus error bars), and the posterior process with a
        hard box constraint added (cyan, mean plus error bars). The
        continuous time potential is defined $(2x_t)^8 I_{[1/2,2/3]}(t)$
        which describe soft box constraints and we assume two hard box
        discrete likelihood terms $I_{[-0.25,0.25]}(x_{t_1})$ and
        $I_{[-0.25,0.25]}(x_{t_2})$ placed at $t_1=1/3$ and $t_2=2/3$. The
        prior is defined by the parameters $a_t=-1$, $c_t=4 \pi \cos(4\pi
        t)$ and $b_t=4$. Right: The purely Gaussian approximation of EP
        fails to anticipate the skew of the distribution, which a correction
        to EP in the style of (\ref{eq:correc_marginal}) is able to capture.
        Figure extracted from ~\citealp{Cseke:AILDP}.}
    \end{center}
\end{figure}

We can apply our EP approximation if we discretize in time, 
\begin{equation}
    p(\{y_{t_i}^d\},\{y_t^c\} | \{x_t\}) \propto \prod_{t_i \in T_d}
    p(y_{t_i}^d|x_{t_i}) \times \prod_k \exp\lbrace-\Delta_{t_k}
    V(t_k,y_t^c,x^{t_k}) \rbrace \;.  \end{equation}
An important question: Does EP remains meaningful in the limit $\Delta t
\rightarrow 0$?  It can be shown that EP continues to be applicable in this
limit for smooth approximations. In figure \ref{fig:33} we have a process
with a drift, with a hard box constraint that is softened in order to apply
the approximation.

\section{Conclusion: open problems with EP}
We have seen how EP applies across a broad range  of theoretical and
practical inference problems, we can conclude with some open problems.

A major open problem with EP is scaling for structured approximation.
Applications like TrueSkill can work with independent variable
approximations, but structured approximations should be more powerful.
Another challenge is creating versions of the algorithm that parallelize.

Convergence properties of EP related approximations are not well understood.
In certain approximation algorithms, or in special applications, there are
non-rigorous arguments that connect non-convergence of fast methods with
weakness of the underlying model assumptions, but this needs to be
generalized.

The free energies are not known to be bounds except for a few special cases.
Bounds are important since they would allow proofs for convergence of
various schemes, and allow certain systematic extensions of approximations.  

It would be interesting to understand performance bounds based on very
general criteria: one such framework with which a connection might be made
are PAC-Bayes bounds.

\section{Acknowledgements}
The authors thank Elizabeth Harrison for providing an audio recording of the
lecture. We also thank the organizers of the school for coordinating our
contribution.

AM was supported by FAPESP under grant 13/01213-8.

\bibliographystyle{plainnat}
\addcontentsline{toc}{section}{References}
\bibliography{main}

\end{document}